\documentclass[twocolumn]{template}

\usepackage[utf8]{inputenc}             
\usepackage[T1]{fontenc}                
\usepackage{url}                        
\usepackage{booktabs}                   
\usepackage{multirow}
\usepackage{colortbl}
\usepackage{multicol}
\usepackage{amsfonts}                   
\usepackage{nicefrac}                   
\usepackage{microtype}                  
\usepackage[dvipsnames]{xcolor}         

\usepackage{latexsym}

\usepackage{graphicx}
\usepackage{float}
\usepackage{subcaption}
\usepackage{wrapfig}
\usepackage{lipsum}
\usepackage{makecell}
\usepackage[table]{xcolor}
\usepackage[most]{tcolorbox}
\usepackage{enumitem}
\usepackage{parskip}
\usepackage{tikz}
\usepackage{amsmath}
\usepackage{booktabs}
\usepackage{multirow}
\usepackage{tabularx}
\usepackage{colortbl}
\usepackage{subcaption}
\usepackage[export]{adjustbox} 
\definecolor{myblue}{RGB}{230,245,255}
\definecolor{mygreen}{RGB}{1,109,101}
\definecolor{mydarkblue}{rgb}{0.68, 0.85, 1.0}
\definecolor{mydarkblue2}{rgb}{0,0.08,0.45}
\definecolor{mydarkblue3}{RGB}{151,204,255}
\definecolor{cvprblue}{rgb}{0.21,0.49,0.74}
\definecolor{oxfordblue}{RGB}{0,33,71}
\definecolor{oxfordroyalblue}{RGB}{29,66,166}
\usepackage{pifont}
\newcommand{\cmark}{\ding{51}}
\usepackage{amssymb}
\usepackage{stfloats}
\usepackage{bm}

\usepackage{tabularx} 
\usepackage{ragged2e} 
\newcolumntype{L}{>{\RaggedRight\hangafter=1\hangindent=0em}X}

\usepackage{enumitem}

\usepackage{amsmath}
\usepackage{amssymb}
\usepackage{mathtools}
\usepackage{amsthm}

\setboolean{logo}{true}    

\usepackage[linesnumbered,ruled,vlined]{algorithm2e}

\hypersetup{
    colorlinks=true,
    linkcolor=red,
    citecolor=Cerulean,
    filecolor=magenta,      
    urlcolor=magenta,
}

\usepackage[capitalize,noabbrev]{cleveref}
\crefname{section}{§}{§§}
\Crefname{section}{§}{§§}

\usepackage{calligra}
\DeclareMathAlphabet{\mathcalligra}{T1}{calligra}{m}{n}

\usepackage{pifont}

\theoremstyle{plain}

\theoremstyle{definition}

\theoremstyle{remark}

\renewcommand{\paragraph}[1]{\vspace{1mm}\noindent\textbf{#1}}

\DeclareCaptionLabelFormat{cont}{#1~#2\alph{ContinuedFloat}}
\usepackage[most]{tcolorbox}
\tcbset{
  promptbox/.style={
    top=10pt,
    colback=lightgray!20,
    colframe=Black,
    colbacktitle=NavyBlue,
    enhanced,
    center,
    attach boxed title to top center={yshift=-0.1in,xshift=0.0in},
    boxed title style={boxrule=0pt,colframe=white,},
  }
}
\newtcolorbox{promptbox}[2][]{promptbox, title=#2,#1}
\tcbset{
  takeawaybox/.style={
    top=10pt,
    colback=lightgray!20,
    colframe=Black,
    colbacktitle=BurntOrange,
    enhanced,
    center,
    attach boxed title to top center={yshift=-0.1in,xshift=0.0in},
    boxed title style={boxrule=0pt,colframe=white,},
  }
}
\newtcolorbox{takeawaybox}[2][]{takeawaybox, title=#2,#1}
\tcbset{
  observationbox/.style={
    top=10pt,
    colback=lightgray!20,
    colframe=Black,
    colbacktitle=YellowGreen,
    enhanced,
    center,
    attach boxed title to top center={yshift=-0.1in,xshift=0.0in},
    boxed title style={boxrule=0pt,colframe=white,},
  }
}
\newtcolorbox{observationbox}[2][]{observationbox, title=#2,#1}

\usepackage{xspace}

\newcommand\blfootnote[1]{%
  \begingroup
  \renewcommand\thefootnote{}\footnote{#1}%
  \addtocounter{footnote}{-1}%
  \endgroup
}

\usepackage{CJK}

\renewcommand{\topfraction}{0.9}
\renewcommand{\bottomfraction}{0.8}
\renewcommand{\textfraction}{0.1}
\renewcommand{\floatpagefraction}{0.85}

\title{ViCuR: Visual Cues as Recoverable Privilege for Multimodal On-Policy Distillation}

  \author[1,2]{Kanghui Tian\textsuperscript{*}}
  \author[3]{Siyuan Liu\textsuperscript{*}}                                                                                  
  \author[1]{Ziang Yan}
  \author[3]{Sheng Xia}                                                                                                      
  \author[2]{Shuai Dong}
  \author[1]{Yi Wang\textsuperscript{$\dagger$}}

  \affil[1]{Shanghai AI Laboratory}
  \affil[2]{Fudan University}
  \affil[3]{Nanjing University}

  \begin{abstract}
On-policy distillation (OPD) improves reasoning by training a student on trajectories sampled from its own policy under supervision from a teacher. In multimodal reasoning, a common extension is to use a \emph{privileged teacher} that observes training-time-only signals such as reference answers or rationales. However, such answer-side privilege creates a train-test mismatch: the teacher's supervision may depend on signals unavailable to the student, encouraging shortcut imitation rather than visually grounded reasoning. We propose \textbf{ViCuR}, a visually grounded privileged-teacher distillation framework that replaces answer-side privilege with \emph{visual cues} (query-related evidence in the input). Because these cues are derived from the same visual input available at inference, their evidence is recoverable by the student. To support this, ViCuR introduces a lightweight \emph{cue recovery module} that uses dedicated sink-token cross-attention during prefill to aggregate task-relevant visual evidence into an internal representation, without changing the inference interface or requiring auxiliary cue-generation losses.
Across seven benchmarks with Qwen3-VL-2B and 8B students, ViCuR consistently improves over answer-based on-policy self-distillation by +1.19 and +1.24 on overall average performance. It also extends naturally to stronger-teacher OPD, surpassing OPD baselines by +0.64 and +1.08, with consistent out-of-domain gains at the 8B scale. These results show that, in multimodal on-policy distillation, the design of teacher privilege is as important as teacher strength.
\end{abstract}

\begin{document}

\maketitle
\blfootnote{$\dagger$ Corresponding author}
\blfootnote{$*$ Equal contribution}
\blfootnote{Code is at \url{https://github.com/tiankanghui/ViCuR}}
\vspace{-3.7em}
\section{Introduction}
\label{sec:intro}

Multimodal reasoning tasks such as geometry problem solving~\cite{chen-etal-2021-geoqa,he-etal-2024-olympiadbench}, chart reasoning~\cite{masry-etal-2022-chartqa,yue2023mmmu}, and visual question answering~\cite{vqa2,wang2025internvideo25empoweringvideomllms,dong2025interleaved} require models to integrate vision with multi-step reasoning. Improving such capabilities through on-policy learning is appealing because supervision is applied on trajectories induced by the student's own policy, directly targeting the states and errors the model encounters at deployment. On-policy distillation (OPD) has therefore emerged as a practical way to improve reasoning: a teacher policy supervises student-sampled trajectories, helping the student learn better generation behavior under its own rollout distribution. 

Teacher advantage in OPD can arise from two different sources. In the standard setting, teacher and student observe the same input, and the teacher is stronger. In a more general \emph{privileged-teacher} setting, the teacher is additionally conditioned on training-only information unavailable to the student at inference. This setting is common in on-policy self-distillation (OPSD), where teacher and student share the same backbone but the teacher is strengthened with extra signals such as reference answers, verified rationales, or environment feedback~\cite{OPSD,sdpo}. While privileged teachers can provide stronger supervision, they also raise a design challenge: \emph{what kind of privileged information is helpful for distillation in multimodal reasoning?}

\begin{figure*}[t]
\setlength{\abovecaptionskip}{5pt}  
\setlength{\belowcaptionskip}{-10pt}  
\centering
\includegraphics[width=1.0\linewidth]{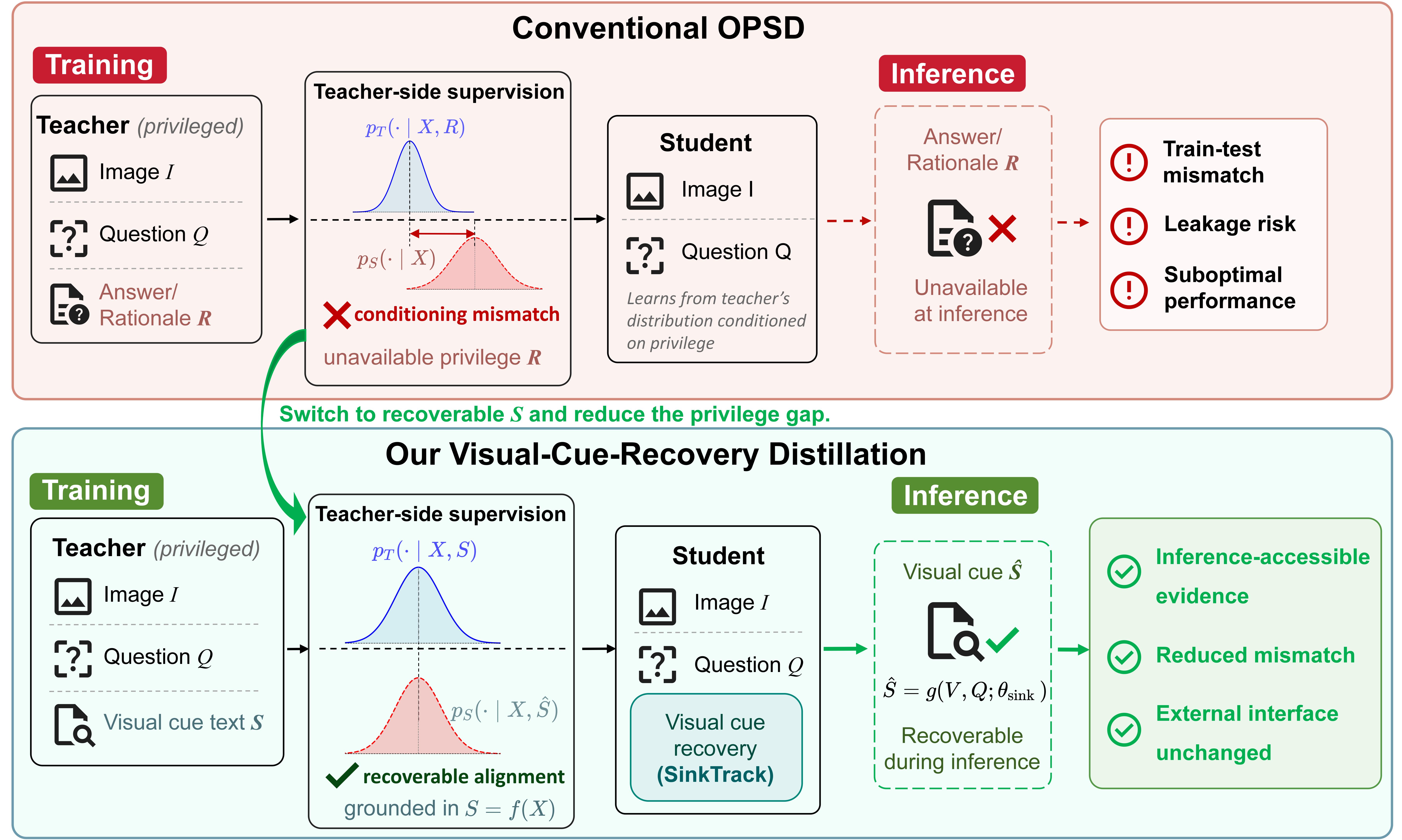}
\caption{Conventional OPSD conditions teacher supervision on answer/rationale privilege \(R\), inducing a train--test mismatch. Our ViCuR instead uses visually grounded cues \(S\), allowing the student to recover corresponding evidence internally while preserving the standard inference interface.}
\label{fig:placeholder}
\end{figure*}

Existing multimodal OPSD methods commonly use answer- or rationale-based privilege~\cite{lin2026visd,vision-opd}. Concretely, given standard inference input $z=(x_v,x_q)$ consisting of visual input $x_v$ and question $x_q$, the student acts as $\pi_S(\cdot \mid z)$, while the teacher is conditioned on additional privileged signal $p$, yielding $\pi_T(\cdot \mid z, p)$. When $p$ contains reference answers or rationales, teacher supervision depends on answer-side information that is unreachable at inference, creating a \emph{privilege-induced train-test mismatch}: the student may fit answer-aware patterns rather than learning reasoning grounded in visual evidence~\cite{yang2026selfdistilledrlvr}. This issue is particularly problematic in multimodal reasoning, where correctness should be supported by evidence in the visuals, such as geometric relations in diagrams, or salient events in videos. In other words, the most useful privilege for multimodal reasoning should strengthen supervision while remaining aligned with information the student can access at inference.

In this work, we propose \textbf{ViCuR} (\textbf{Vi}sual \textbf{Cu}e \textbf{R}ecovery), a visually grounded privileged-teacher distillation framework. It replaces answer-based privileges with \emph{visual cues}: question-relevant evidence grounded in the input image or video. Although the cue text itself is provided only to the teacher during training, its underlying source remains accessible to the student through the standard visual-question interface. This changes the role of privilege from revealing answer-side information to highlighting inference-recoverable evidence, mitigating the privileged-information mismatch. In conventional answer-based distillation, the teacher policy $\pi_T(\cdot \mid z, p)$ depends on privileged variables $p$ that are not recoverable from the student's input $z$. By contrast, when the privilege is a visual cue $c$ derived from the visual input and question, $c \sim p(c \mid x_v, x_q)$, the teacher is conditioned on evidence whose source is already contained in $z$. ViCuR thus does not eliminate privilege altogether, but replaces inaccessible answer privilege with recoverable visual privilege (\S\ref{subsec:recoverable_privilege}).

Replacing the privilege type introduces a new challenge: the student does not observe cues explicitly, so it must learn to exploit cue-relevant evidence from visual tokens. To address this, we introduce a lightweight \emph{cue recovery module}. Inspired by the attention-sink mechanism~\cite{streaming_attn_sink,liu2026sinktrack}, we equip a designated sink token with dedicated cross-attention parameters at selected transformer layers; during prefill, this token aggregates task-relevant information from visual tokens into an internal cue-level representation. The module is trained end-to-end under the distillation objective alone, preserving the standard inference interface with no autoregressive overhead (\S\ref{subsec:sinktrack}).

Though our primary focus is privileged-teacher self-distillation, teacher advantage from model scale and from privilege design are orthogonal, so ViCuR extends naturally to stronger-teacher OPD. We evaluate on seven multimodal reasoning benchmarks using Qwen3-VL-2B and 8B students. ViCuR consistently outperforms answer-based OPSD, improving overall average performance by +1.19 and +1.24, with especially strong gains on in-domain and near-domain mathematical reasoning tasks. It also extends effectively to stronger-teacher OPD, where it improves over OPD baselines by +0.64 and +1.08 and delivers consistent out-of-domain gains at the 8B scale.

Our main contributions are as follows:
\begin{itemize}[leftmargin=*]
    \item We identify the bottleneck of existing multimodal OPSD: answer-based privileges induce a train-test mismatch that can encourage shortcut imitation rather than visually grounded reasoning.
    \item We propose \textbf{ViCuR}, which replaces answer-side privilege with visually grounded cues and introduces a lightweight sink-token cross-attention module for internal cue recovery, without changing the inference interface or requiring auxiliary objectives.
    \item We show that ViCuR consistently improves over answer-based OPSD across seven benchmarks and extends naturally to stronger-teacher OPD, where it further benefits from teacher scaling.
\end{itemize}

\section{Related Work}
\label{sec:related_work}

\paragraph{On-Policy Distillation and Self-Distillation.}
Knowledge distillation~\cite{Hinton2015DistillingTK,kang2023knowledge-aug,shu2025llava-mod} transfers reasoning capabilities from a large teacher to a compact student.
For reasoning tasks, recent work extends distillation from offline imitation of teacher-generated traces to OPD, where the teacher supervises trajectories sampled from the student itself, improving alignment with the student's inference-time distribution~\cite{ICLR2024_opd,lu2025onpolicy-thinkinglab,minillm,video_opd}. Standard OPD typically requires a stronger external teacher.
OPSD removes this requirement by using a single backbone for both teacher and student, instead strengthening the teacher with privileged conditioning such as verified reasoning traces~\cite{OPSD} or feedback-derived supervision~\cite{sdpo,vpd}.
While OPSD avoids the cost of a separate teacher, its effectiveness depends on the design of the privileged signal, which we discuss in the context of privileged information below.

\paragraph{Privileged Information for Distillation.}
Learning with privileged information~\cite{vapnik2015learning-privileged-info} augments training with signals that are unavailable at test time.
In reasoning distillation, common privilege types include gold answers, rationales, verifier scores, and environment feedback~\cite{hsieh2023distillstep,OPSD,sdpo}. Recent self-distillation work has begun exploring answer-agnostic alternatives, such as conciseness instructions~\cite{opsdc}, context-conditioned prompts or historical traces~\cite{opcd}, and evidence-centered regional crops~\cite{vision-opd}.
However, when privileges are answer-dependent, they can cause \emph{privileged information leakage}~\cite{yang2026selfdistilledrlvr}: formally showed that the teacher's supervision may depend on variables the student cannot access at inference time, encouraging the student to fit answer-aware patterns rather than learn genuine reasoning.

\paragraph{Attention Sinks.}
Attention sinks refer to the phenomenon that special tokens such as \texttt{<BOS>} attract disproportionately high attention mass across transformer layers~\cite{streaming_attn_sink}, effectively serving as persistent information anchors~\cite{li2024snapkv}.
SinkTrack~\cite{liu2026sinktrack} leverages this property by using the sink token to aggregate and retain contextual information during long-sequence generation, improving context fidelity without architectural changes.
\section{Method}
\label{sec:method}

We study multimodal on-policy distillation in the \emph{privileged-teacher} regime, where the teacher accesses additional training-time information unavailable to the student at inference. Our method, \textbf{ViCuR}, is built on a simple principle: privileged information is most useful when grounded in evidence that remains accessible at test time. We first formalize the privilege-induced mismatch and explain why visual cues mitigate it (\S\ref{subsec:problem_setup}), then present the ViCuR architecture and training objective (\S\ref{subsec:sinktrack}--\S\ref{subsec:implementation}).

\subsection{Problem Formulation}
\label{subsec:problem_setup}

Consider a multimodal reasoning task with visual input $I$, question $Q$, and answer $y$. Let $X = (I,Q)$ denote the standard multimodal input available at both training and inference. In OPD, the student samples a trajectory $\hat{y} \sim p_\theta(\cdot \mid X)$,
and the teacher provides token-level supervision on this student-generated trajectory. We consider the \emph{privileged-teacher} setting, in which the teacher is additionally conditioned on a training-time-only variable \(p\), yielding
$p_T(y \mid X,p)$,
while the student uses only $X$. This formulation subsumes both standard same-input OPD, where $p=\emptyset$, and privileged distillation settings, where $p$ may contain extra side information. In existing multimodal OPSD, this privilege is commonly answer- or rationale-based; in ours, it is instead a visually grounded cue.

\paragraph{Privilege-Induced Gap in Answer-Based Distillation.}
\label{subsec:privilege_gap}
In OPSD, the teacher is conditioned on an answer- or rationale-based privilege $R$, producing $p_T(y \mid X,R)$. Because \(R\) is unavailable at inference time, the teacher's next-token distribution may depend on information the student cannot access during deployment. Following~\cite{yang2026selfdistilledrlvr}, we quantify this mismatch by the conditional mutual information as $I(Y_t;\,R \mid X, Y_{<t})$.
When $R$ contains answer-dependent information, such as a gold answer or reference rationale, this quantity is generally nonzero. Distillation then asks the student to imitate a target policy whose behavior is not fully determined by the input $X$, creating a privilege-induced train-test mismatch. In multimodal reasoning, this mismatch is problematic because it encourages the student to fit answer-aware supervision patterns or shortcuts rather than learning to ground its reasoning in visuals.

\paragraph{Visual Cues as Recoverable Privilege.}
\label{subsec:recoverable_privilege}
We replace answer-based privilege $R$ with a visually grounded cue $S$: a description of question-relevant evidence, such as geometric relations or key events. Although the student does not receive cue text explicitly at inference, the evidence from which $S$ is derived remains present in the standard input $X$.

For analysis, we idealize cue construction as a deterministic mapping $S = f(X)$, where $f$ extracts cue text from the inference-time input. Under this abstraction, $H(S \mid X)=0$, so conditioning on $S$ introduces no additional information beyond $X$:
\begin{equation}
I(Y_t;\,S \mid X, Y_{<t}) = 0.
\end{equation}
Thus, the cue-conditioned teacher distribution
\begin{equation}
p_T(\cdot \mid X,S,Y_{<t})
=
p_T(\cdot \mid X,f(X),Y_{<t})
\end{equation}
depends only on information determined by the inference-time multimodal input.

We formalize this as \textbf{Proposition~1} (Appendix~\ref{app:proof_prop1}): under the deterministic abstraction $S{=}f(X)$, visually grounded cues do not introduce the privilege-induced conditional information gap associated with answer-dependent variables, as the resulting teacher supervision depends only on information determined by the inference-time input $X$.

In practice, cue text may be generated by an external model or annotator, so $S$ need not be strictly deterministic given $X$, yielding $H(S \mid X) > 0$. However, because visual cues describe evidence visible in the input rather than answer-dependent information, the inaccessible component of cue-based privilege is substantially smaller than that of answer-based privilege. The deterministic abstraction should therefore be interpreted as an idealized limit: visual cues \emph{reduce} the privilege gap rather than eliminating it entirely, and we show empirically that this reduction is sufficient to yield consistent gains.

Proposition~1 does not require the student to reconstruct cue text explicitly. Instead, it reframes the remaining challenge as \emph{evidence recovery}: the relevant information is present in $X$, but the student must learn to extract and represent it internally.

\subsection{ViCuR: Sink-Based Cue Recovery}
\label{subsec:sinktrack}

To address this challenge, we introduce \textbf{ViCuR} (\textbf{Vi}sual \textbf{Cu}e \textbf{R}ecovery), a lightweight student-side module that recovers cue-relevant evidence from visual representations without changing the standard multimodal inference interface. A naive alternative would be to append cue text to the student input or require the student to generate cues before answering. However, this would alter the inference pipeline and add generation overhead. We instead seek a mechanism that operates within the student's existing architecture.

Our framework uses an asymmetric teacher-student design. The teacher is conditioned on the visual input, question, and cue text, producing \(p_T(y \mid I,Q,S)\). The student predicts with \(p_\theta(y \mid I,Q)\), augmented with a cue recovery module.

\begin{figure}
\setlength{\abovecaptionskip}{5pt}  
\setlength{\belowcaptionskip}{-10pt}  
\centering
\includegraphics[width=1.0\linewidth]{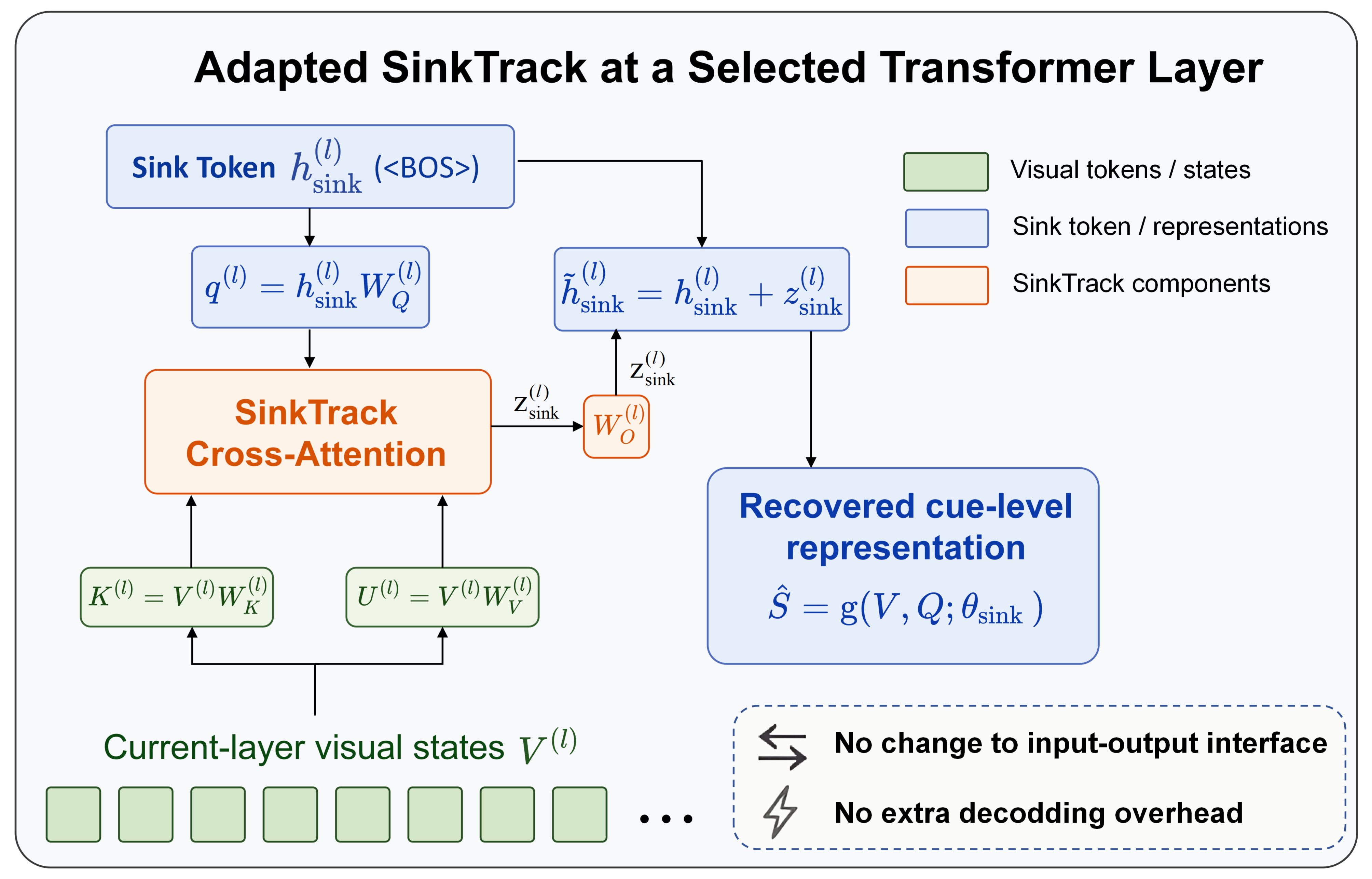}
\caption{Cue recovery module at a selected transformer layer. The sink token aggregates visual states via dedicated cross-attention during prefill, producing cue representations without autoregressive overhead.}
\label{fig:sinktrack}
\end{figure}

Let \(V^{(0)}=\phi(I) \in \mathbb{R}^{m \times d}\) be the visual token sequence from the visual encoder. At transformer layer \(\ell\), let \(V^{(\ell)} \in \mathbb{R}^{m \times d}\) denote the current visual representations, and let \(h^{(\ell)}_{\mathrm{sink}} \in \mathbb{R}^{d}\) denote the hidden state of a designated sink token (e.g., \texttt{<BOS>}). As shown in Fig.~\ref{fig:sinktrack}, at selected layers during prefill, the sink token queries the visual tokens through a dedicated cross-attention branch:
\begin{equation}
\normalsize
\begin{aligned}
q^{(\ell)} &= h^{(\ell)}_{\mathrm{sink}} W_Q^{(\ell)}, \ 
K^{(\ell)} = V^{(\ell)} W_K^{(\ell)}, \\
U^{(\ell)} &= V^{(\ell)} W_V^{(\ell)}, \\
z^{(\ell)}_{\mathrm{sink}} &= \mathrm{Attn}\bigl(q^{(\ell)}, K^{(\ell)}, U^{(\ell)}\bigr) W_O^{(\ell)},
\end{aligned}
\end{equation}
where \(W_Q^{(\ell)}, W_K^{(\ell)}, W_V^{(\ell)}, W_O^{(\ell)}\) are recovery-module-specific parameters (we use \(U^{(\ell)}\) for the value projection to avoid conflict with visual tokens \(V^{(\ell)}\)). The aggregated evidence is added back to the sink state via a residual connection:
\begin{equation}
\tilde{h}^{(\ell)}_{\mathrm{sink}}
=
h^{(\ell)}_{\mathrm{sink}} + z^{(\ell)}_{\mathrm{sink}}.
\end{equation}

Because the sink token appears at the beginning of the sequence, its updated state is visible to all subsequent question and answer tokens through standard causal attention. We denote the resulting internal evidence representation as
\begin{equation}
    \hat{S} = g(V,Q;\theta_{\mathrm{sink}}),
\end{equation}
where \(\theta_{\mathrm{sink}}\) collects all recovery-module parameters. The symbol \(\hat{S}\) refers to an internal hidden-state representation, not generated cue text. Although the cross-attention equations above involve only \(h^{(\ell)}_{\mathrm{sink}}\) and \(V^{(\ell)}\), the question \(Q\) influences \(\hat{S}\) implicitly through optimization: gradients from the question-dependent prediction loss guide \(\theta_{\mathrm{sink}}\) to selectively aggregate visual features from \(V^{(\ell)}\) that are most relevant for answering \(Q\).

\paragraph{Why Dedicated Cross-Attention.}
We introduce \emph{dedicated} cross-attention parameters, which are free to learn question-relevant visual selection without interfering with the pretrained self-attention pathway and can absorb the implicit supervision signal from the distillation objective. This differs from the original SinkTrack mechanism~\cite{liu2026sinktrack}, which reuses the existing attention computation for generic context anchoring.

\subsection{Training Objective}
\label{subsec:theory_gradient}

The cue recovery module is trained jointly with the student under the on-policy distillation objective, without any auxiliary cue-generation or text-matching loss. For a student-sampled trajectory $\hat{y} \sim p_\theta(\cdot \mid X)$,
the teacher evaluates each token with $p_T(\hat{y}_n \mid X,S,\hat{y}_{<n})$,
while the student assigns $p_\theta(\hat{y}_n \mid X,\hat{S},\hat{y}_{<n})$.
Following sampled-token on-policy distillation~\cite{li2026rethinking}, we define the token-level advantage
\begin{align}
A_n(X,\hat{y}) &= \log p_T(\hat{y}_n \mid X,S,\hat{y}_{<n}) \nonumber \\
&- \log p_\theta(\hat{y}_n \mid X,\hat{S},\hat{y}_{<n}),
\end{align}
which is treated as a stop-gradient constant, as in standard policy-gradient optimization~\cite{lu2025onpolicy-thinkinglab,jin2026eopd}. The distillation surrogate loss is
\begin{align}
&\mathcal{L}_{\mathrm{d}}(\theta)
= -\mathbb{E}_{X \sim \mathcal{D}}
\mathbb{E}_{\hat{y} \sim p_\theta(\cdot \mid X)}
\biggl[\\
&\frac{1}{|\hat{y}|}
\sum_{n=1}^{|\hat{y}|} 
\mathrm{sg}\!\bigl[A_n(X,\hat{y})\bigr] \nonumber 
\times
\log p_\theta(\hat{y}_n \mid X,\hat{S},\hat{y}_{<n})
\biggr],
\label{eq:distill_loss}
\end{align}
where \(\mathrm{sg}[\cdot]\) denotes stop-gradient. Here we use a PPO-style policy loss~\cite{schulman2017proximal} with importance ratios and clipping~\cite{luo2026demystifying}; the expression above isolates the gradient pathway through which the recovery module is trained.

Since \(\hat{S}=g(V,Q;\theta_{\mathrm{sink}})\), gradients reach the recovery module through
\begin{align}
&\nabla_{\theta_{\mathrm{sink}}}\mathcal{L}_{\mathrm{d}}
= -\mathbb{E}\biggl[
\frac{1}{|\hat{y}|}\sum_{n=1}^{|\hat{y}|}
\mathrm{sg}[A_n]\\
&\nabla_{\hat{S}}
\log p_\theta(\hat{y}_n \mid X,\hat{S},\hat{y}_{<n}) \nonumber \cdot
\nabla_{\theta_{\mathrm{sink}}} g(V,Q;\theta_{\mathrm{sink}})
\biggr].
\label{eq:grad_sink}
\end{align}
This expression shows that the recovery module is updated when its representation \(\hat{S}\) affects the student's prediction on teacher-evaluated tokens. Whenever the teacher assigns higher probability than the student to the sampled token, the gradient encourages the student to move its representation in a direction that better supports teacher-aligned behavior; when the opposite holds, the gradient suppresses unhelpful representations. In this way, the distillation objective implicitly encourages the sink-based module to retain task-relevant visual evidence aligned with the teacher's cue-conditioned supervision, even though the student never observes cue text directly.

A detailed analysis of these gradient dynamics is provided in Appendix~\ref{app:analysis_prop2}; a full-vocabulary KL counterpart is discussed in Appendix~\ref{app:kl_counterpart}.

\subsection{Implementation Details}
\label{subsec:implementation}

We train the recovery module jointly with the student under the distillation objective described above. The recovery module is inserted every five transformer layers and uses the full current-layer visual token sequence as keys and values. It runs only during prefill and adds no autoregressive decoding overhead, introducing only a modest parameter overhead relative to the base model.

\section{Experiments}
\label{sec:experiments}

We evaluate ViCuR to answer three questions: (1)~Does replacing answer-based privilege with visual cues improve over conventional distillation baselines, including both same-backbone OPSD and stronger-teacher OPD? (2)~Do the gains generalize across in-domain, near-domain, and out-of-domain benchmarks? (3)~Which components of ViCuR matter most, and how does the method behave as teacher strength increases?

\paragraph{Data.}
We train on the Vision R1 training set~\cite{visionr1}, from which teacher-side visual cue text is constructed. Evaluation covers three generalization levels: Vision R1-Test~\cite{visionr1} (in-domain); DynaMath~\cite{zou2025dynamath}, MathVista~\cite{lu2024mathvista}, WeMath~\cite{wemath}, and MathVerse~\cite{mathverse} (near-domain mathematical reasoning); and MMMU-Val~\cite{yue2023mmmu} with Video-MME~\cite{videomme} (out-of-domain multimodal understanding). Geometry3K~\cite{geo3k} is reserved for ablation studies due to its smaller scale and diagram-centric nature.

\paragraph{Models and Baselines.}
We use Qwen3-VL-2B-Instruct (2B) and Qwen3-VL-8B-Instruct (8B)~\cite{bai2025qwen3vltechnicalreport} as student backbones. In OPSD~\cite{OPSD}, the teacher shares the student's backbone; in OPD~\cite{lu2025onpolicy-thinkinglab}, the teacher is larger (Qwen3-VL-8B for 2B, Qwen3-VL-32B for 8B). We compare: base model, GRPO~\cite{shao2024deepseekmathpushinglimitsmathematical} (reward-based, no teacher), OPSD, OPSD+ViCuR, OPD, and OPD+ViCuR. Our primary comparisons are against OPSD and OPD, isolating the effect of visual-cue privilege and cue recovery. Details are in Appendix~\ref{app:implementation_details}.

\subsection{Main Results}
\label{subsec:main_results}

Tab.~\ref{tab:main_results} reports results across all benchmarks for both 2B and 8B students. We compare ViCuR against the corresponding distillation baseline under the same teacher configuration, and then discuss how the gains vary across generalization regimes.

\begin{table*}[t]
\setlength{\abovecaptionskip}{5pt}
\setlength{\belowcaptionskip}{-10pt}
\centering
\small
\setlength{\tabcolsep}{3pt}
\renewcommand{\arraystretch}{1.35}

\newcommand{\valdiff}[2]{%
  #1{\scriptsize #2}%
}

\setlength{\aboverulesep}{0pt}
\setlength{\belowrulesep}{0pt}
\resizebox{\textwidth}{!}{
\begin{tabular}{lcccccccc}
\toprule
\multirow{2}{*}{\textbf{Method}}
& \multicolumn{1}{c}{\textbf{In-Domain}}
& \multicolumn{4}{c}{\textbf{Near-Domain}}
& \multicolumn{2}{c}{\textbf{Out-of-Domain}}
& \multicolumn{1}{c}{\textbf{Overall}} \\
\cmidrule(lr){2-2} \cmidrule(lr){3-6} \cmidrule(lr){7-8} \cmidrule(lr){9-9}
& \textbf{Vision R1-Test}
& \textbf{DynaMath}
& \textbf{MathVista}
& \textbf{WeMath}
& \textbf{MathVerse}
& \textbf{MMMU-Val}
& \textbf{Video-MME}
& \textbf{Avg.} \\
\midrule
\multicolumn{9}{l}{\textit{\textbf{Qwen3-VL-2B-Instruct}}} \\
Base Model           & 23.90 & 53.23 & 61.0  & 31.43 & 39.09 & 51.11 & 58.2    & 45.42 \\
GRPO                 & 28.17 & 55.49 & 60.8  & 40.19 & 44.90 & 52.00 & 58.0    & 48.51 \\
OPSD                 & 25.00 & 52.37 & 61.2  & 30.57 & 37.56 & 49.56 & 57.9    & 44.88 \\
\rowcolor{myblue}
\textbf{OPSD+ViCuR} & \valdiff{26.58}{\textcolor{blue}{(+1.58)}} & \valdiff{51.60}{\textcolor{red}{(-0.77)}} & \valdiff{63.4}{\textcolor{blue}{(+2.2)}} & \valdiff{33.90}{\textcolor{blue}{(+3.33)}} & \valdiff{40.00}{\textcolor{blue}{(+2.44)}} & \valdiff{49.22}{\textcolor{red}{(-0.34)}} & \valdiff{57.8}{\textcolor{red}{(-0.1)}} & \valdiff{46.07}{\textcolor{blue}{(+1.19)}} \\
OPD                  & 26.34 & 52.87 & 64.4  & 35.71 & 40.56 & 51.00 & 57.6    & 46.93 \\
\rowcolor{myblue}
\textbf{OPD+ViCuR}  & \valdiff{26.83}{\textcolor{blue}{(+0.49)}} & \valdiff{52.38}{\textcolor{red}{(-0.49)}} & \valdiff{63.6}{\textcolor{red}{(-0.8)}} & \valdiff{36.00}{\textcolor{blue}{(+0.29)}} & \valdiff{42.31}{\textcolor{blue}{(+1.75)}} & \valdiff{53.30}{\textcolor{blue}{(+2.30)}} & \valdiff{58.6}{\textcolor{blue}{(+1.0)}} & \valdiff{47.57}{\textcolor{blue}{(+0.64)}} \\
\midrule
\multicolumn{9}{l}{\textit{\textbf{Qwen3-VL-8B-Instruct}}} \\
Base Model           & 41.22 & 67.13 & 75.8  & 55.33 & 53.40 & 66.56 & 65.9    & 60.76 \\
GRPO                 & 58.75 & 67.44 & 76.6  & 67.05 & 56.22 & 65.56 & 65.8    & 65.35 \\
OPSD                 & 42.20 & 65.99 & 73.5  & 43.62 & 52.92 & 62.33 & 66.5    & 58.15 \\
\rowcolor{myblue}
\textbf{OPSD+ViCuR} & \valdiff{45.00}{\textcolor{blue}{(+2.80)}} & \valdiff{67.17}{\textcolor{blue}{(+1.18)}} & \valdiff{74.1}{\textcolor{blue}{(+0.6)}} & \valdiff{46.29}{\textcolor{blue}{(+2.67)}} & \valdiff{53.25}{\textcolor{blue}{(+0.33)}} & \valdiff{63.33}{\textcolor{blue}{(+1.00)}} & \valdiff{66.6}{\textcolor{blue}{(+0.1)}}  & \valdiff{59.39}{\textcolor{blue}{(+1.24)}} \\
OPD                  & 51.09 & 69.22 & 78.1  & 58.38 & 60.05 & 64.33 & 66.0 & 63.88 \\
\rowcolor{myblue}
  \textbf{OPD+ViCuR}  & \valdiff{52.07}{\textcolor{blue}{(+0.98)}} & \valdiff{70.70}{\textcolor{blue}{(+1.48)}} & \valdiff{76.7}{\textcolor{red}{(-1.4)}} & \valdiff{60.57}{\textcolor{blue}{(+2.19)}} & \valdiff{61.55}{\textcolor{blue}{(+1.50)}} & \valdiff{66.56}{\textcolor{blue}{(+2.23)}} & \valdiff{66.6}{\textcolor{blue}{(+0.6)}} & \valdiff{64.96}{\textcolor{blue}{(+1.08)}} \\
\bottomrule
\end{tabular}
}
\caption{Main results across in-domain, near-domain, and out-of-domain benchmarks. ViCuR consistently improves the overall average over the corresponding distillation baseline (OPSD or OPD) at both 2B and 8B scales. Deltas in parentheses are relative to the corresponding baseline without ViCuR.}
\label{tab:main_results}
\end{table*}

\paragraph{ViCuR Improves Over Answer-based OPSD.}
Under same-backbone self-distillation, OPSD+ViCuR raises the average by +1.19 (2B) and +1.24 (8B) over conventional answer-based OPSD. The gains are strongest on benchmarks with clear visual grounding demands, such as WeMath (+3.33 at 2B, +2.67 at 8B), supporting our hypothesis that visually grounded cues yield more useful supervision than answer-side privilege.

Notably, conventional OPSD underperforms the base model at both scales (44.88 vs.\ 45.42 for 2B; 58.15 vs.\ 60.76 for 8B), consistent with prior analyses of answer-conditioned self-distillation~\cite{yang2026selfdistilledrlvr}. ViCuR mitigates this degradation, surpassing the base model at 2B (46.07 vs.\ 45.42) and substantially narrowing the gap at 8B. A few benchmarks show small decreases at 2B scale (DynaMath -0.77, Video-MME -0.1), which we attribute to limited student capacity for tasks requiring broader numerical or temporal reasoning beyond the geometry-heavy training distribution; these drops disappear or reverse at 8B, supporting this interpretation.

\paragraph{ViCuR Improves Stronger-Teacher OPD.}
ViCuR extends to stronger-teacher OPD, improving the average by +0.64 at 2B and +1.08 at 8B. Gains are broadly distributed across five of seven (2B) and six of seven (8B) benchmarks. MathVista is the exception, with small drops at both scales (-0.8 and -1.4); analysis shows ViCuR improves geometry-heavy subsets but is slightly weaker on computation-dominated subsets (arithmetic reasoning, math word problems), where the bottleneck is numerical processing rather than visual grounding.

\paragraph{ViCuR Improves Transfer Beyond Training Domain.}
The strongest out-of-domain improvements appear in the OPD setting. In particular, OPD + ViCuR improves MMMU-Val by +2.30 at 2B and +2.23 at 8B, and improves Video-MME at both scales. This suggests reducing answer-side mismatch helps stronger teachers transfer more effectively beyond the mathematical reasoning training distribution, improving broader multimodal understanding and video reasoning as well.

\subsection{Ablation Studies}
\label{subsec:ablation}

We next analyze which components of ViCuR drive the gains and how the method behaves under different teacher-student configurations.

\paragraph{Component Analysis.}
\label{abl:main_component}
Table~\ref{tab:main_component_analysis} isolates the contributions of visual cues and the cue recovery module in the OPSD setting on Geometry3K. 
Replacing answer-based privilege with visual cues yields the largest individual improvement: \textbf{+2.80} for the 2B student and \textbf{+4.36} for the 8B student. This indicates that the dominant benefit comes from redesigning the teacher-side privilege. By contrast, adding the cue recovery module alone produces limited gains (+1.17 for 2B and +0.09 for 8B), as expected, because the student-side module is still trained against answer-conditioned teacher supervision in this variant. The full combination achieves the best results at both scales: \textbf{+4.80} for 2B and \textbf{+4.65} for 8B. This confirms that the two components are complementary: visual cues improve the quality of teacher supervision and cue recovery helps the student internalize the corresponding evidence.

The recovery module adds 100.7M parameters for the 2B model (4.52\%) and 536.9M for the 8B model (5.77\%). Despite this, it operates only during prefill and introduces no autoregressive decoding overhead: per-token decoding speed is unaffected, and wall-clock training time remains comparable to standard OPSD (Appendix~\ref{app:computational_overhead} provides detailed training and inference timing). To verify that the trend is not specific to Geometry3K, we further evaluate on geometry subsets of MathVista and DynaMath, observing similar patterns. Detailed results are in Appendix~\ref{app:ablation_main_component}.

\begin{table}[t]
\setlength{\abovecaptionskip}{5pt}
\setlength{\belowcaptionskip}{-10pt}
\centering
\small 
\renewcommand{\arraystretch}{1.2} 

\begin{tabularx}{\columnwidth}{l
>{\centering\arraybackslash}X
>{\centering\arraybackslash}X
>{\centering\arraybackslash}X
>{\centering\arraybackslash}X}
\toprule
\multirow{2}{*}{\textbf{Method}} & \multicolumn{2}{c}{\textbf{2B Student}} & \multicolumn{2}{c}{\textbf{8B Student}} \\
\cmidrule(lr){2-3} \cmidrule(lr){4-5}
& Accuracy & $\Delta$ & Accuracy & $\Delta$ \\
\midrule
Baseline         & 34.14 & --    & 52.40 & --    \\
w/ visual cue    & 36.94 & +2.80 & 56.76    & +4.36    \\
w/ cue recovery  & 35.31 & +1.17 & 52.49   & +0.09    \\
\rowcolor{myblue}
\textbf{ViCuR  }          & \textbf{38.94} & \textbf{+4.80} & \textbf{57.05} & \textbf{+4.65} \\
\bottomrule
\end{tabularx}

\caption{Component analysis under the OPSD setting on Geometry3K-test. Accuracy is reported as the average over the final 100 training steps. $\Delta$ denotes the absolute improvement over the corresponding baseline.}
\label{tab:main_component_analysis}
\end{table}

\begin{figure*}[!tb]
\setlength{\abovecaptionskip}{5pt}
\setlength{\belowcaptionskip}{-10pt}
\centering

\begin{subfigure}[t]{0.49\textwidth}
    \centering
    \begin{tikzpicture}
        \node[anchor=south west, inner sep=0] (image) at (0,0)
            {\includegraphics[width=\linewidth]{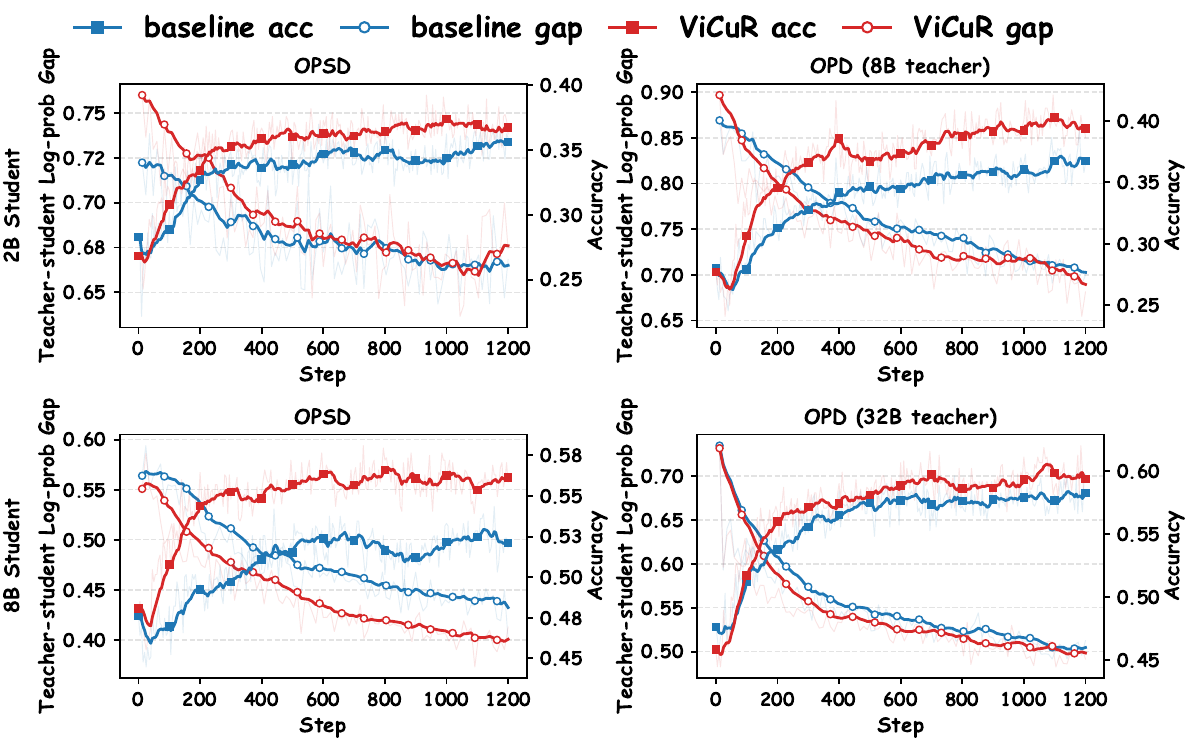}};
        \node[anchor=north west, inner sep=0pt, xshift=0pt, yshift=-1pt]
            at (image.north west) {\scriptsize\textbf{(a)}};
    \end{tikzpicture}
    \captionsetup{labelformat=empty}
    \caption{}
    \label{fig:logprob_overall}
\end{subfigure}
\hfill
\begin{subfigure}[t]{0.49\textwidth}
    \centering
    \begin{tikzpicture}
        \node[anchor=south west, inner sep=0] (image) at (0,0)
            {\includegraphics[width=\linewidth]{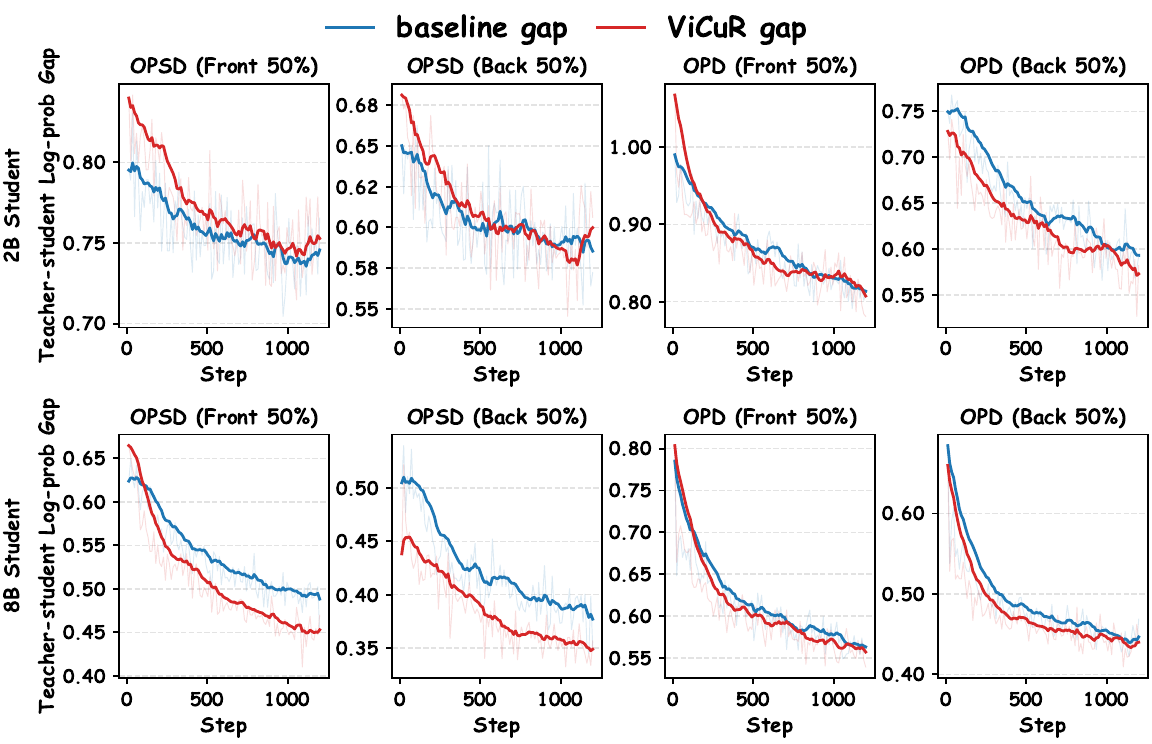}};
        \node[anchor=north west, inner sep=0pt, xshift=0pt, yshift=-4pt]
            at (image.north west) {\scriptsize\textbf{(b)}};
    \end{tikzpicture}
    \captionsetup{labelformat=empty}
    \caption{}
    \label{fig:logprob_stage}
\end{subfigure}

\caption{Training dynamics of student models under different teacher configurations and distillation paradigms.
(a) Overall training dynamics.
(b) Stage-wise training dynamics, where student-generated sequences are split into the front 50\% and back 50\% of tokens. For both OPSD and OPD, we use the answer-based method as the baseline.}
\label{fig:logprob_analysis}
\end{figure*}

\paragraph{Scaling Behavior across Teacher-Student Configurations.}
We further examine how ViCuR behaves as student capacity and teacher strength vary on Geometry3K (Fig.~\ref{fig:scale_analysis} in Appendix~\ref{app:scale_analysis}). ViCuR improves all four teacher-student configurations, with gains ranging from +3.35 to +6.88. The largest gain (+6.88) occurs for the 8B student distilled from a 32B teacher, substantially exceeding the self-distillation gain (+4.65), suggesting visual-cue grounding and teacher scaling are complementary when the student is large enough. For the 2B student, the 8B teacher does not produce a larger gain than self-distillation (+3.35 vs.\ +4.80), indicating that once the student becomes the bottleneck, additional teacher strength is harder to exploit.

\paragraph{Token-Level Log-Probability Dynamics.}
To better understand training behavior, we track the teacher-student token-level log-probability gap together with student test accuracy on Geometry3K throughout training. We report both the full sequence and a stage-wise split into the front and back 50\% of tokens (Fig.~\ref{fig:logprob_analysis}).

Fig.~\ref{fig:logprob_overall} shows that ViCuR often begins with a larger teacher-student gap than answer-based baselines, but later reaches parity or achieves a lower gap while maintaining higher test accuracy under both OPSD and OPD. This indicates that cue-conditioned supervision is not easier to imitate at the outset; instead, it appears to provide better learning signals, ultimately leading to stronger student behavior.

Fig.~\ref{fig:logprob_stage} further shows that both baseline methods and ViCuR have smaller gaps on the back half of the sequence than on the front half, suggesting that student alignment is generally easier near answer-proximal tokens. More importantly, ViCuR consistently achieves a smaller back-half gap than the answer-based baseline. This suggests that grounding the student in relevant visual evidence improves not only early reasoning but also downstream answer-stage alignment. Additional plots are provided in Appendix~\ref{app:ablation_log_prob}.

\begin{figure}[!tbp]
\setlength{\abovecaptionskip}{5pt}
\setlength{\belowcaptionskip}{-10pt}
    \centering
    \includegraphics[width=\linewidth]{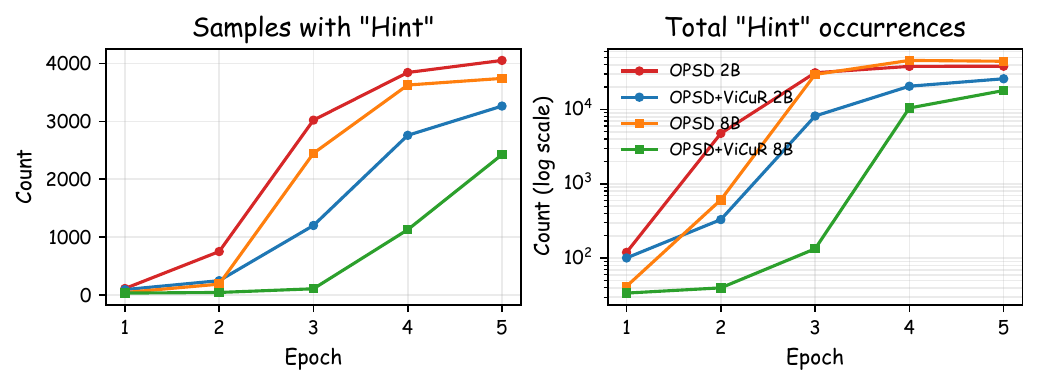}
\caption{Hint leakage under OPSD, measured by affected sample count and total ``Hint'' occurrences across epochs. Lower values indicate less leakage.}
    \label{fig:hint_leakage_stats}
\end{figure}

\paragraph{Hint Leakage in Student Rollouts.}
To probe privilege leakage in student behavior, we count explicit occurrences of ``Hint'' in student rollouts. Following the teacher prompt design of prior work~\cite{OPSD}, privileged teacher-side content is injected using a \texttt{[Hint]} format. Student-side generation of ``Hint'' serves as a direct diagnostic of training-only prompt-pattern reproduction.

Fig.~\ref{fig:hint_leakage_stats} shows that answer-based OPSD exhibits steadily increasing hint leakage over training, whereas ViCuR consistently reduces both the number of affected samples and the total number of ``Hint'' occurrences. Although this lexical diagnostic does not capture all forms of shortcut learning, it provides direct behavioral evidence that visual-cue supervision reduces explicit dependence on teacher-side privileged patterns. Full statistics, including the OPD setting, are provided in Appendix~\ref{app:hint_leakage_stats}.

\paragraph{Qualitative Evidence and Attention Visualization.}
To complement the quantitative results, we provide case studies and SinkTrack cross-attention visualizations in Appendix~\ref{app:qualitative_cases}. These examples show that ViCuR reduces concrete visual grounding errors induced by answer-based OPSD (e.g., incorrect label-segment binding in geometry diagrams) and that the sink-based recovery module aggregates visual tokens consistent with teacher-side cue evidence, supporting the claimed mechanism.
\section{Conclusion}
We presented \textbf{ViCuR}, which replaces answer-based privilege in multimodal on-policy distillation with visually grounded cues and equips the student with a lightweight sink-based recovery module for internal evidence aggregation. Across seven benchmarks and two student scales, ViCuR consistently improves over answer-based OPSD and extends effectively to stronger-teacher OPD, with the strongest gains on visually grounded reasoning tasks. Our results suggest that in multimodal distillation, \emph{designing privilege around inference-recoverable visual evidence is as important as teacher strength itself}.
\section{Limitations}
Our work focuses on improving on-policy multimodal distillation by replacing answer-based privileged information with visually grounded cues and by introducing an internal cue recovery module. While this design reduces the mismatch caused by answer-based privilege, it still depends on the quality of the constructed visual cues during training. If the cue generator misses task-relevant evidence, introduces non-visual information, or describes the visual scene ambiguously, the teacher-side supervision may become less reliable.

In addition, the cue recovery module introduces extra trainable parameters and optimization complexity. The component analysis suggests that its benefit does not increase monotonically across model scales: for the 8B student on Geometry3K, full ViCuR performs slightly worse than the visual-cue-only variant, even though ViCuR still improves over the corresponding OPSD/OPD baselines in the main results. This indicates that larger students may require more careful training of the recovery module, such as more parameter-efficient designs, staged optimization, richer cue recovery data, or additional regularization.

\section*{Acknowledgments}
We thank the Shanghai Artificial Intelligence Laboratory for their institutional support.

\bibliographystyle{plain}
\bibliography{main}


\clearpage
\appendix
\onecolumn

\renewcommand{\topfraction}{0.9}
\renewcommand{\bottomfraction}{0.9}
\renewcommand{\textfraction}{0.1}
\renewcommand{\floatpagefraction}{0.8}
\setcounter{topnumber}{3}
\setcounter{bottomnumber}{3}
\setcounter{totalnumber}{5}

\section{Training and Evaluation Details}
\label{app:implementation_details}

\begin{table}[!htbp]
\centering
\small
\renewcommand{\arraystretch}{1.15}
\setlength{\tabcolsep}{8pt}
\begin{tabular}{lccc}
\toprule
\textbf{Setting} & \textbf{Vision R1 Distillation} & \textbf{Vision R1 GRPO} & \textbf{Geometry3K Ablations} \\
\midrule
Training data & Vision R1 train & Vision R1 train & Geometry3K train \\
Max prompt length & 2048 & 2048 & 1024 \\
Max response length & 4096 & 4096 & 2048 \\
Train batch size & 128 & 128 & 128 \\
Learning rate & \(1\times 10^{-6}\) & \(1\times 10^{-6}\) & \(1\times 10^{-6}\) \\
Rollouts per prompt (\(n\)) & 1 & 5 & 1 \\
GPU allocation & 4 student + 4 teacher & 8 student & 4 student + 4 teacher \\
Training epochs & 5 & 5 & 100 \\
\bottomrule
\end{tabular}
\caption{Key training configurations for the main Vision R1 experiments and the controlled Geometry3K ablations.}
\label{tab:training_details}
\end{table}

All experiments are implemented on top of the VeRL framework~\cite{sheng2025verl}. Unless otherwise specified, experiments are run on a single node with 8 H200 GPUs. For the main Vision R1 experiments, all compared methods are trained on the same training split and evaluated on the same benchmark suite, differing only in the learning objective and supervision source. Geometry3K is used separately for controlled ablation studies. For the Qwen3-VL model family, the first \texttt{<|im\_start|>} token, i.e. the one preceding the system prompt, is selected as the sink token.

\subsection{Training Details}

Tab.~\ref{tab:training_details} summarizes the key training configurations for the main Vision R1 experiments and the controlled Geometry3K ablations.

Unless otherwise specified, all distillation-based experiments use the same sampled-token on-policy distillation framework described in the main text. Teacher-side privileged prompts are provided through the \texttt{teacher\_prompt} field, and images are loaded from the \texttt{images} field. In the distillation-based setting, rollouts are generated with a single student sample per prompt, whereas GRPO uses 5 sampled rollouts per prompt under the same prompt/response length budget.

\subsection{Evaluation Details}

The models are evaluated on \path{MathVista_MINI}, \path{MathVerse_MINI}, 
\path{WeMath}, \path{DynaMath}, \path{MMMU_DEV_VAL}, and 
\path{Video-MME_64frame} with the VLMEvalKit framework~\cite{duan2024vlmevalkit}, following the default dataset definitions in VLMEvalKit. 
Tab.~\ref{tab:eval_details} summarizes the key generation and preprocessing settings during evaluation.

\begin{table}[h!]
\centering
\small
\renewcommand{\arraystretch}{1.1}
\setlength{\tabcolsep}{4pt}
\begin{tabular}{lc}
\toprule
\textbf{Evaluation setting} & \textbf{Value} \\
\midrule
System prompt & \texttt{You are a helpful assistant.} \\
Custom prompt & No \\
Temperature & 0.7 \\
Top-\(p\) & 0.8 \\
Top-\(k\) & 20 \\
Repetition penalty & 1.0 \\
Presence penalty & 1.5 \\
Max new tokens & 4096 \\
Min pixels & 3584 \\
Max pixels & 401408 \\
Min frames & 4 \\
Max frames & 512 \\
\bottomrule
\end{tabular}
\caption{Evaluation hyperparameters used in VLMEvalKit.}
\label{tab:eval_details}
\end{table}

\section{Ablation Details}
\subsection{Main Component Analysis}
\label{app:ablation_main_component}

\begin{table}[!htbp]
\centering
\small
\resizebox{\textwidth}{!}{
\begin{tabular}{llcccccc}
\toprule
\multirow{2}{*}{\textbf{Student}}
& \multirow{2}{*}{\textbf{Setting}}
& \multicolumn{3}{c}{\textbf{Method Components}}
& \multicolumn{2}{c}{\textbf{MathVista}}
& \multicolumn{1}{c}{\textbf{DynaMath}} \\
\cmidrule(lr){3-5} \cmidrule(lr){6-7} \cmidrule(lr){8-8}
& & \textbf{Answer} & \textbf{Visual Cue} & \textbf{Cue Recovery}
& \textbf{Geo. Reas.} & \textbf{Geo. Prob.} & \textbf{Plane Geo.} \\
\midrule

\multirow{9}{*}{2B}
& Base
& & & 
& 66.95 & 69.23 & 50.91 \\

\cmidrule(lr){2-8}

& \multirow{4}{*}{OPSD}
& \cmark & & 
& 67.36 & 68.75 & 51.82 \\
&
& & \cmark & 
& \underline{73.22} & \underline{75.48} & \textbf{54.55} \\
&
& & \cmark & \cmark
& \textbf{74.48} & \textbf{76.92} & \underline{53.77} \\

\cmidrule(lr){2-8}

& \multirow{4}{*}{\makecell[l]{OPD\\(8B teacher)}}
& & & 
& 73.64 & 75.00 & 49.22 \\
&
& \cmark & & 
& \underline{74.06} & 75.48 & 48.96 \\
&
& & \cmark & 
& \textbf{75.73} & \textbf{78.37} & \textbf{54.29} \\
&
& & \cmark & \cmark
& \underline{74.06} & \underline{77.40} & \underline{51.30} \\

\midrule

\multirow{9}{*}{8B}
& Base
& & & 
& 84.10 & 86.54 & 57.66 \\

\cmidrule(lr){2-8}

& \multirow{4}{*}{OPSD}
& \cmark & & 
& 85.36 & \underline{86.54} & 57.03 \\
&
& & \cmark & 
& \textbf{87.03} & \textbf{88.46} & \underline{59.35} \\
&
& & \cmark & \cmark
& \underline{85.77} & \underline{86.54} & \textbf{60.00} \\

\cmidrule(lr){2-8}

& \multirow{4}{*}{\makecell[l]{OPD\\(32B teacher)}}
& & & 
& 84.94 & 86.54 & 61.43 \\
&
& \cmark & & 
& 86.61 & 87.98 & 61.69 \\
&
& & \cmark & 
& \textbf{88.28} & \textbf{90.38} & \underline{61.82} \\
&
& & \cmark & \cmark
& \underline{87.87} & \underline{89.90} & \textbf{63.12} \\

\bottomrule
\end{tabular}
}
\vspace{0.5em}
\caption{Additional cross-benchmark ablation results. All models are trained on Geometry3K and evaluated on geometry-related subsets of MathVista and DynaMath. We report the \texttt{geometry reasoning} and \texttt{geometry problem solving} subsets for MathVista and the \texttt{plane geometry} subset for DynaMath. The solid-geometry subset of DynaMath is omitted because Geometry3K contains little solid-geometry supervision. Checkmarks indicate whether each method uses answer-based privileged supervision, visual-cue-based supervision, or cue recovery. Rows without checkmarks under OPD denote the vanilla OPD setting. Bold and underlined numbers indicate the best and second-best results within each student-setting group, respectively.}
\label{tab:cross_dataset_ablation}
\end{table}

We provide additional ablation results under a cross-benchmark evaluation protocol. All models are trained on Geometry3K and evaluated on geometry-related subsets of MathVista and DynaMath. This setting is more challenging than the in-domain Geometry3K evaluation in Sec.~\ref{abl:main_component} because the target benchmarks differ in problem style, annotation format, and visual distribution. As shown in Tab.~\ref{tab:cross_dataset_ablation}, two findings stand out:

(1) \textbf{Visual-cue supervision consistently improves transfer performance over the answer-based baseline.} For example, in the 2B OPSD setting, replacing answer with visual cues improves MathVista geometry reasoning from 67.36 to 73.22, MathVista geometry problem solving from 68.75 to 75.48, and DynaMath plane geometry from 51.82 to 54.55. Similar gains are also observed under cross-scale OPD and with the 8B student. These results indicate that the benefit of visual cues is not limited to the Geometry3K test set, but also transfers to related geometry benchmarks.

(2) \textbf{Cue recovery shows greater sensitivity under cross-benchmark evaluation.}
Although cue recovery brings additional gains in several cases, such as 2B OPSD on the two MathVista subsets and 8B OPSD/OPD on DynaMath plane geometry, it does not uniformly outperform visual-cue supervision alone. 
One possible reason is that the recovery module introduces more extra capacity for larger students: it adds 536.9M parameters to the 8B student, accounting for 5.77\% of the model, compared with 100.7M parameters and 4.52\% for the 2B student. 
With the same amount of training data, this larger module may overfit source-domain visual-cue patterns. 
When target benchmarks differ in visual style, reasoning format, or cue granularity, the recovered cues can become less reliable, explaining the mixed gains of cue recovery across target subsets.

\subsection{Scaling Behavior across Teacher-Student Configurations}
\label{app:scale_analysis}

\begin{figure}[!htbp]
\setlength{\belowcaptionskip}{-10pt}
    \centering
    \includegraphics[width=0.75\linewidth]{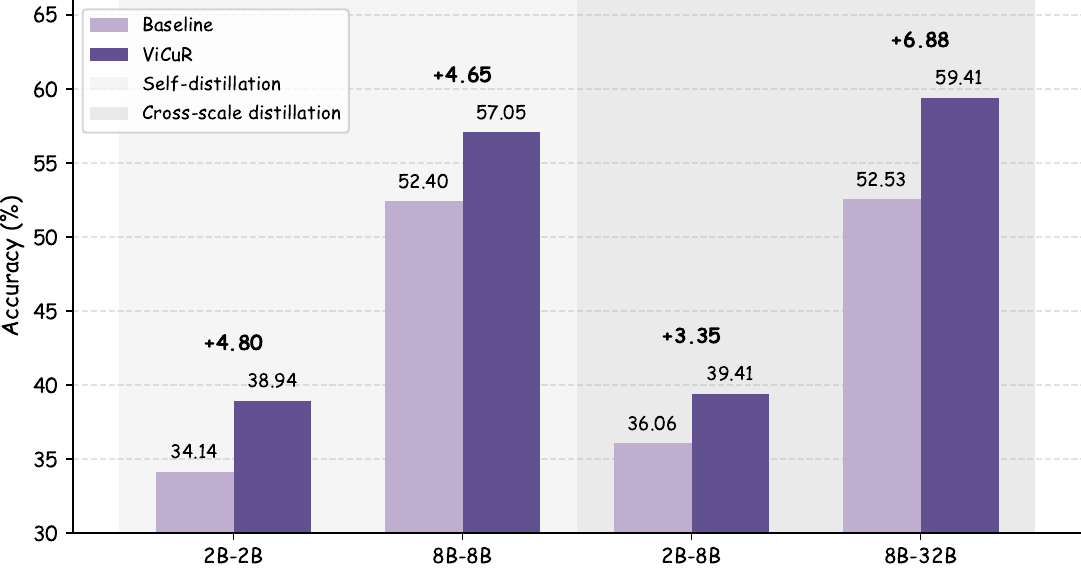}
    \caption{ViCuR gains across teacher--student configurations on Geometry3K. Each x-axis label denotes a student--teacher pair. Baselines are answer-based OPSD and vanilla OPD, respectively.}
    \label{fig:scale_analysis}
\end{figure}

Fig.~\ref{fig:scale_analysis} examines how ViCuR behaves as both student capacity and teacher strength vary. ViCuR improves all four configurations, with the largest gain (+6.88) for the 8B student distilled from a 32B teacher. For the 2B student, the stronger 8B teacher does not outperform self-distillation (+3.35 vs.\ +4.80), indicating that once the student becomes the bottleneck, additional teacher strength is harder to exploit. Overall, these results suggest that ViCuR scales most favorably when both the teacher and the student are sufficiently capable.

\subsection{Token-level Log-probability Analysis}
\label{app:ablation_log_prob}
\begin{figure}[!htbp]
    \centering
    \includegraphics[width=\textwidth]{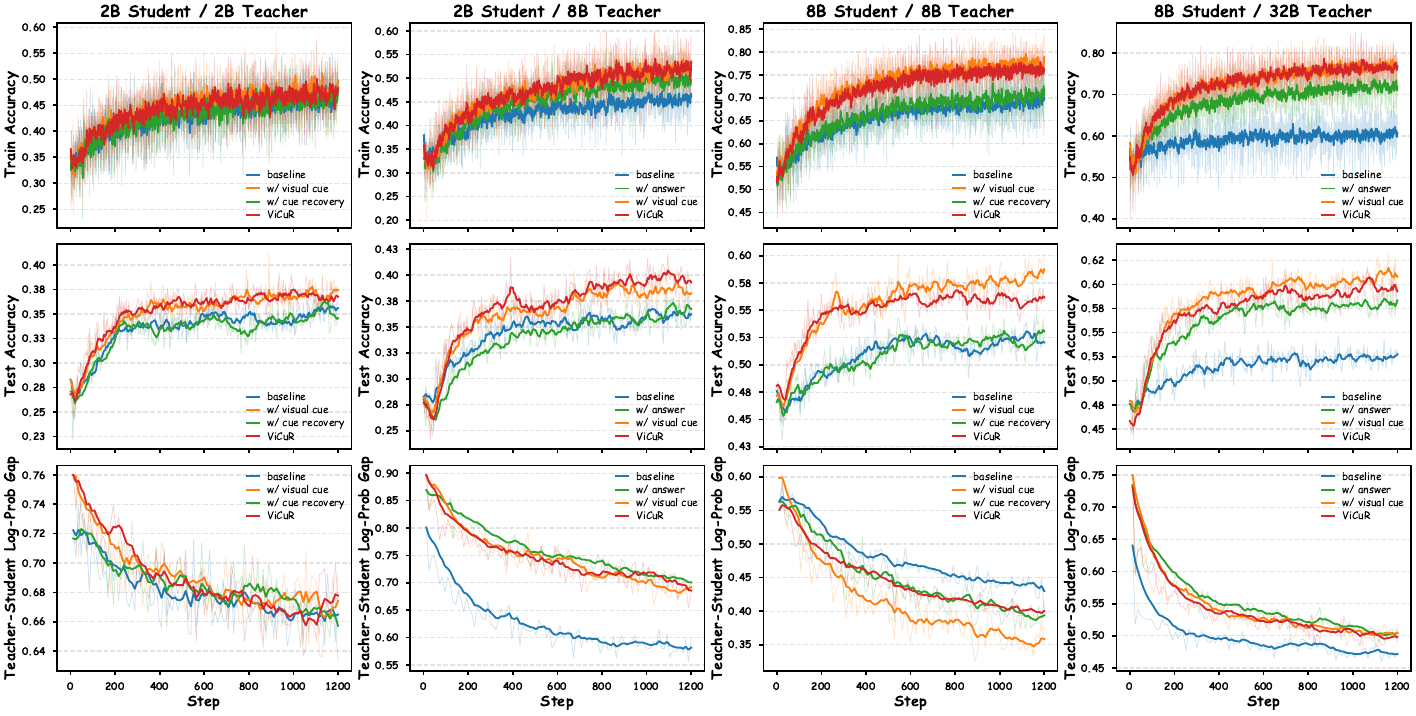}
    \caption{
    Token-level analysis on Geometry3K during training.
    We report the student's train and test accuracy and the teacher--student token-level log-probability gap under different teacher--student configurations. For OPSD, we use the answer-based method as the baseline, and for OPD, we adopt vanilla OPD without any additional input information as the baseline.
    }
    \label{fig:logprob_analysis_app}
\end{figure}

\paragraph{Full-rollout log-probability analysis.}
Fig.~\ref{fig:logprob_analysis_app} tracks the student's accuracy on the train and test splits of Geometry3K and the teacher--student token-level log-probability gap on the full rollout trajectory during training under different teacher--student configurations. 
In the 2B-student/2B-teacher setting, the answer-based baseline obtains a smaller gap but much lower accuracy, suggesting that easier token-level imitation does not necessarily lead to better multimodal reasoning. By contrast, visual cues and ViCuR sometimes maintain a larger gap while achieving higher accuracy, implying that the teacher provides more challenging but more useful supervision grounded in visual evidence. These results support our hypothesis that, for multimodal reasoning, helping the student learn question-relevant visual cues is more important than simply imitating answer-conditioned trajectories.

\begin{figure}[!htbp]
    \centering
    \includegraphics[width=\textwidth]{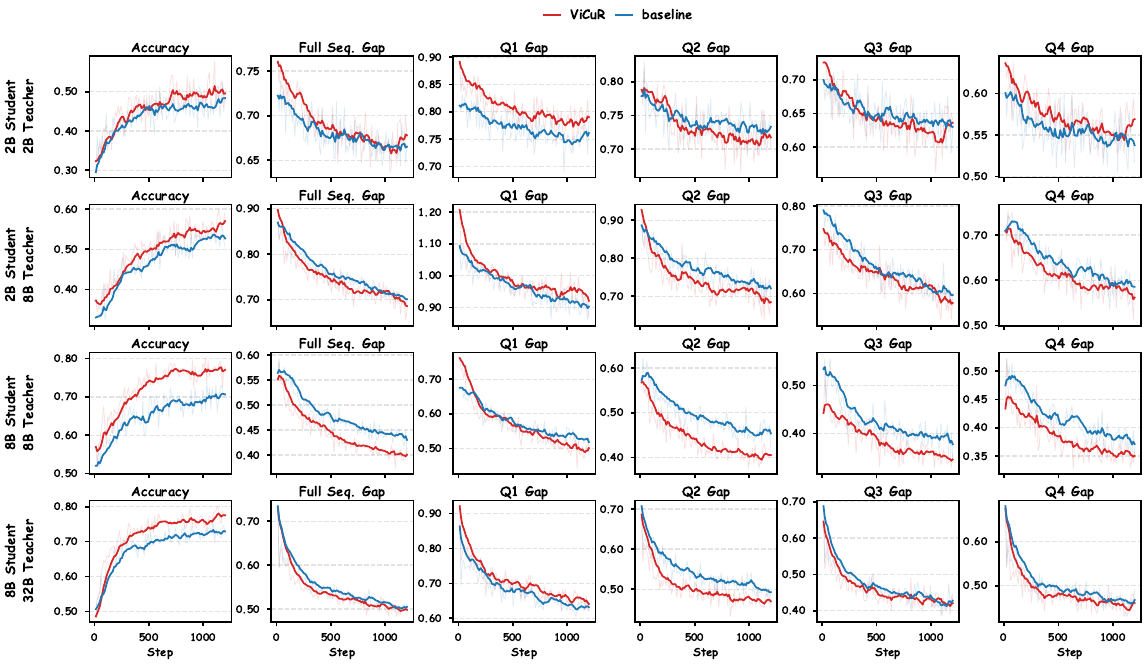}
    \caption{
    Segment-level token-level analysis on Geometry3K.
    We compare ViCuR with the answer-based baseline in terms of validation accuracy, full-sequence teacher--student log-probability gap, and segment-wise gaps over four equal rollout segments. For both OPSD and OPD, we use the answer-based method as the baseline.
    }
    \label{fig:logprob_quarter}
\end{figure}

\paragraph{Segment-level log-probability analysis.}
We further divide each student rollout into four equal segments and analyze the teacher--student token-level log-probability gap in each segment (Fig.~\ref{fig:logprob_quarter}). Compared with the baseline, ViCuR often obtains a comparable or smaller gap, especially in the stronger-student settings, suggesting that visual-cue-based supervision does not weaken token-level alignment. The segment-level results further show that the benefit of ViCuR is not confined to the final answer tokens. The gap of ViCuR is consistently smaller than that of the baseline across the middle two segments of the rollout, where the model typically performs visual reasoning and answer derivation. Although ViCuR can exhibit a larger gap in the early segment, this is consistent with the additional difficulty of grounding the reasoning process in question-relevant visual evidence rather than directly imitating answer-conditioned supervision. Overall, these results indicate that visual cues provide effective supervision throughout the reasoning trajectory, helping the student learn to ground its reasoning in question-relevant visual evidence rather than fitting answer-aware patterns.

\subsection{Computational Overhead Analysis}
\label{app:computational_overhead}

The cue recovery module adds 100.7M parameters for the 2B model (4.52\%) and 536.9M for the 8B model (5.77\%). It operates only during prefill and introduces no autoregressive decoding overhead. We quantify the wall-clock cost below.

\paragraph{Training time.}
Tab.~\ref{tab:training_time} reports the total training time on the Vision R1 dataset (5 epochs, 275 steps; single node, 8$\times$H200). ViCuR introduces only modest overhead over the corresponding baseline at both scales, confirming that the recovery module does not substantially increase training cost.

\begin{table}[h]
    \centering
    \small
    \renewcommand{\arraystretch}{1.15}
    \setlength{\tabcolsep}{6pt}
    \begin{tabular}{lcc}
    \toprule
    \textbf{Method} & \textbf{2B} & \textbf{8B} \\
    \midrule
    OPSD & 6:37:03 & 9:13:53 \\
    OPSD + ViCuR & 6:47:52 & 8:18:38 \\
    OPD & 6:53:41 & 9:07:18 \\
    OPD + ViCuR & 6:57:21 & 8:43:16 \\
    \bottomrule
    \end{tabular}
    \caption{Total wall-clock training time (hh:mm:ss) on Vision R1 (5 epochs, 275 steps; 8$\times$H200).}
    \label{tab:training_time}
    \end{table}

\paragraph{Inference time.}
Tab.~\ref{tab:inference_time} compares inference latency with and without the cue recovery module on a single H200 GPU (bfloat16, flash attention 2, max generation 2,048 tokens). We test two input conditions representing typical low- and high-resolution multimodal inputs: a small-image input (88 tokens total, 64 visual tokens) and a large-image input (1,821 tokens total, 1,782 visual tokens). All measurements are averaged over 10 runs after 3 warmup iterations.

\begin{table}[!htbp]
\centering
\small
\renewcommand{\arraystretch}{1.15}
\setlength{\tabcolsep}{5pt}
\resizebox{\textwidth}{!}{
\begin{tabular}{llcccccc}
\toprule
\textbf{Input} & \textbf{Model} & \textbf{Input len} & \textbf{Vis.\ tokens} & \textbf{Prefill (ms)} & \textbf{Decoding (ms/tok)} & \textbf{Avg output len} & \textbf{Total gen (ms)} \\
\midrule
\multirow{4}{*}{\makecell[l]{Small image}}
& 2B baseline & 88 & 64 & 30.64 & 18.31 & 633 & 11,625 \\
& 2B + ViCuR & 88 & 64 & 38.14 \textcolor{gray}{\scriptsize(+24.5\%)} & 16.78 & 207 & 3,506 \\
& 8B baseline & 88 & 64 & 36.92 & 24.14 & 165 & 4,019 \\
& 8B + ViCuR & 88 & 64 & 44.06 \textcolor{gray}{\scriptsize(+19.3\%)} & 21.47 & 211 & 4,583 \\
\midrule
\multirow{4}{*}{\makecell[l]{Large image}}
& 2B baseline & 1,821 & 1,782 & 68.79 & 18.18 & 369 & 6,778 \\
& 2B + ViCuR & 1,821 & 1,782 & 73.60 \textcolor{gray}{\scriptsize(+7.0\%)} & 16.68 & 428 & 7,213 \\
& 8B baseline & 1,821 & 1,782 & 129.20 & 23.36 & 255 & 6,078 \\
& 8B + ViCuR & 1,821 & 1,782 & 134.71 \textcolor{gray}{\scriptsize(+4.3\%)} & 21.55 & 446 & 9,745 \\
\bottomrule
\end{tabular}
}
\caption{Inference latency on a single H200 GPU. We report prefill time, per-token decoding time, average output length, and total generation time under small-image (88 tokens, 64 visual) and large-image (1,821 tokens, 1,782 visual) inputs. Total generation time varies primarily due to different output lengths across runs; per-token decoding time is the comparable metric.}
\label{tab:inference_time}
\end{table}

Two observations support the computational efficiency of ViCuR. First, per-token decoding time is unaffected by the recovery module, as it is inactive during autoregressive generation. Across all configurations, the per-token decoding time remains comparable between baseline and ViCuR (differences are within run-to-run variance due to different generated content lengths). Since decoding dominates total inference time---for example, in the 2B large-image case, decoding accounts for $>$98\% of total generation time---the recovery module has negligible impact on end-to-end inference cost.
Second, the prefill overhead introduced by the recovery module scales sub-linearly with visual token count. On the small-image input (64 visual tokens), the relative overhead is $\sim$20\%; on the large-image input (1,782 visual tokens), it drops to 4--7\%. This is because the cross-attention computation in the recovery module uses a single query (the sink token) against visual keys/values, yielding $O(m)$ cost where $m$ is the number of visual tokens---a negligible addition compared to the $O(n^2)$ self-attention over the full sequence during prefill. In the high-resolution settings common in multimodal reasoning, this overhead becomes increasingly marginal.

\section{Qualitative Comparison of Teacher-Side Hints}
\label{app:hint_examples}

\begin{figure}[!htbp]
\centering

\begin{tcolorbox}[colback=white,colframe=black!60,title={Case 1: Intersecting chords},fonttitle=\bfseries]
\begin{minipage}[t]{0.17\textwidth}
    \centering
    \vspace{0pt}
    \includegraphics[width=0.68\linewidth]{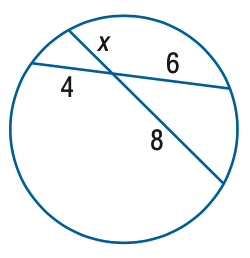}

    \vspace{0.5em}
    \textbf{Question:} Find \(x\).

    \vspace{0.3em}
    \textbf{Answer:} \(3\)
\end{minipage}
\hfill
\begin{minipage}[t]{0.79\textwidth}
    \vspace{0pt}
    \begin{tcolorbox}[colback=red!3,colframe=red!65!black,title={Reasoning hint},fonttitle=\bfseries]
    \small
    \texttt{[Hint]: The problem is solved by applying the Intersecting Chords Theorem, which equates the product of the segment lengths of one chord ($x \cdot 8$) to the product of the segment lengths of the other chord ($4 \cdot 6$).}
    \end{tcolorbox}

    \vspace{-0.5em}

    \begin{tcolorbox}[colback=blue!3,colframe=blue!65!black,title={Visual cue hint (ours)},fonttitle=\bfseries]
    \small
    \texttt{[Hint]: The diagram shows a circle with two intersecting chords. One chord is divided into segments labeled \(x\) and \(8\), while the other chord is divided into segments labeled \(4\) and \(6\). The intersection lies inside the circle, forming four chord segments with the given lengths.}
    \end{tcolorbox}
\end{minipage}
\end{tcolorbox}

\vspace{-0.8em}

\begin{tcolorbox}[colback=white,colframe=black!60,title={Case 2: Trigonometry in isosceles triangles},fonttitle=\bfseries]
\begin{minipage}[t]{0.17\textwidth}
    \centering
    \vspace{0pt}
    \includegraphics[width=0.68\linewidth]{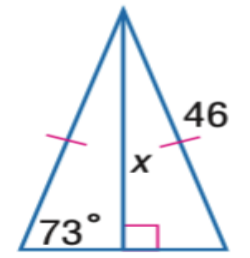}

    \vspace{0.5em}
    \textbf{Question:} Find $x$. Round to the nearest tenth.

    \vspace{0.3em}
    \textbf{Answer:} \(44.0\)
\end{minipage}
\hfill
\begin{minipage}[t]{0.79\textwidth}
    \vspace{0pt}
    \begin{tcolorbox}[colback=red!3,colframe=red!65!black,title={Reasoning hint},fonttitle=\bfseries]
    \small
    \texttt{[Hint]: The problem involves an isosceles triangle where the altitude creates a right triangle with a known hypotenuse (due to congruent sides) and a known base angle, allowing the use of the sine ratio $\left(\sin = \frac{\text{opp}}{\text{hyp}}\right)$ to solve for the height $x$.}
    \end{tcolorbox}

    \vspace{-0.5em}

    \begin{tcolorbox}[colback=blue!3,colframe=blue!65!black,title={Visual cue hint (ours)},fonttitle=\bfseries]
    \small
    \texttt{[Hint]: The diagram shows an isosceles triangle with congruent legs marked by tick marks, each labeled as $46$. A vertical altitude is drawn from the apex to the base, forming a right angle at the base and splitting the triangle into two right triangles. On the left right triangle, the hypotenuse is $46$, one angle is labeled $73^\circ$, and the altitude is labeled $x$, representing the side opposite the $73^\circ$ angle.}
    \end{tcolorbox}
\end{minipage}
\end{tcolorbox}

\caption{Qualitative comparison of teacher-side privileged hints under the same student-facing query. Reasoning hints directly expose the key theorem or solving idea, whereas the visual cue hints used in our framework describe only the question-relevant visual structure grounded in the image.}
\label{fig:hint_examples}
\end{figure}

Fig.~\ref{fig:hint_examples} compares two forms of teacher-side privileged hints given the same student-facing query: one is the reasoning hint that directly exposes the key theorem or solving idea, and the other is the visual cue hint that describes only the question-relevant visual structure grounded in the image.

\section{Hint Leakage Statistics in Student Rollouts}
\label{app:hint_leakage_stats}

We report the complete hint-leakage statistics in Tab.~\ref{tab:hint_leakage_full}. The statistics are computed from student rollout outputs only: we scan the \texttt{output} field, match the complete word ``Hint'' case-insensitively, and aggregate both the number of samples containing at least one match and the total number of matches. Each epoch contains 55 rollout steps. In OPSD, the baseline teacher is conditioned on answer-side privileged hints, whereas ViCuR replaces these with visual cues. In OPD, the baseline does not use privileged hints, while OPD + ViCuR adds teacher-side visual cues. Thus, student-side occurrences of ``Hint'' should be interpreted as a lexical proxy for explicit reproduction of privileged prompt formats, with the OPD baseline serving as a no-privilege reference.

\begin{table}[!htbp]
\centering
\small
\setlength{\tabcolsep}{6pt}
\renewcommand{\arraystretch}{1.15}
\begin{tabular}{@{} l c c c c c @{}}
\toprule
\textbf{Student \& Method} & \multicolumn{5}{c}{\textbf{Epoch}} \\
\cmidrule(lr){2-6}
 & \textbf{1} & \textbf{2} & \textbf{3} & \textbf{4} & \textbf{5} \\
\midrule
\multicolumn{6}{@{}l@{}}{\textbf{2B Student}} \\
OPSD                    & 107/120     & 747/4,768   & 3,021/31,143 & 3,845/37,951 & 4,054/38,151 \\
OPSD + ViCuR            & 88/101      & 243/331     & 1,198/8,164  & 2,758/20,533 & 3,264/25,861 \\
OPD (8B teacher)        & 61/74       & 56/63       & 60/65        & 56/63        & 39/46        \\
OPD + ViCuR (8B teacher)& 56/60       & 81/85       & 123/140      & 172/192      & 247/316      \\
\midrule
\multicolumn{6}{@{}l@{}}{\textbf{8B Student}} \\
OPSD                     & 38/42      & 183/610     & 2,447/29,576 & 3,628/45,629 & 3,743/44,453 \\
OPSD + ViCuR             & 28/34      & 39/40       & 103/134      & 1,125/10,462 & 2,424/17,976 \\
OPD (32B teacher)        & 26/29      & 22/22       & 14/15        & 16/16        & 16/16        \\
OPD + ViCuR (32B teacher)& 25/27      & 40/45       & 38/38        & 32/32        & 44/47        \\
\bottomrule
\end{tabular}
\caption{Complete hint-leakage statistics in student rollout outputs. Each cell reports \textbf{samples / total occurrences}. All configurations are trained with \textbf{Steps = 55} per epoch.}
\label{tab:hint_leakage_full}
\end{table}

\begin{figure}[!htbp]
    \centering
    \includegraphics[width=0.85\linewidth]{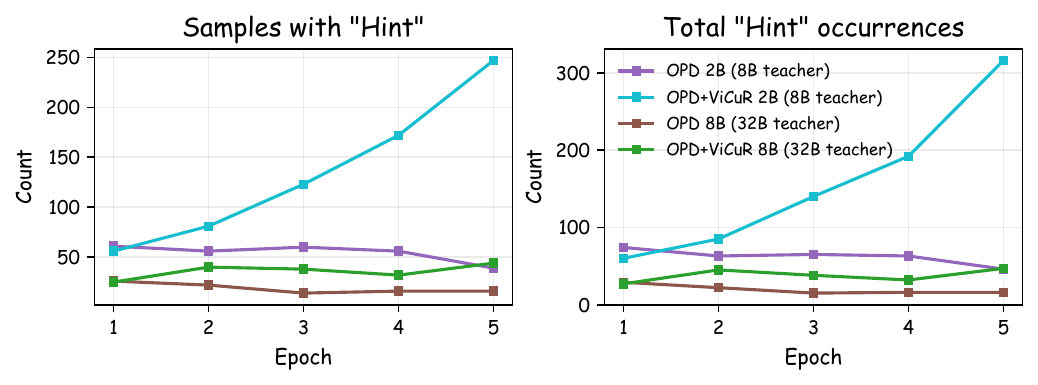}
    \caption{Hint leakage in student rollout outputs under OPD. The OPD baseline does not use teacher-side privileged hints, while OPD + ViCuR introduces teacher-side visual cues. The counts remain low in both settings, showing that visual-cue privilege does not induce the large-scale explicit hint reproduction observed in answer-based OPSD.}
    \label{fig:hint_leakage_opd_stats}
\end{figure}

The OPSD baselines show a sharp increase in explicit hint reproduction, especially in later epochs, indicating that answer-side privileged hints can be copied into student rollouts. ViCuR substantially reduces this effect under OPSD at both student scales. The OPD baseline remains low, as expected, because it does not include privileged hints. More importantly, OPD + ViCuR also stays at a low absolute level despite using teacher-side visual cues; even when its lexical count is slightly higher than OPD, the magnitude is far below the leakage observed in answer-based OPSD. This suggests that visual-cue privilege does not induce the same large-scale explicit hint reproduction as answer-based self-distillation.

\section{Qualitative Case Studies and Attention Visualization}
\label{app:qualitative_cases}

We provide qualitative case studies and SinkTrack cross-attention visualizations using the checkpoints from the main experiments. This appendix complements the main results by examining the central mechanism underlying by ViCuR: replacing answer-side privilege with visually grounded cues is useful only if the student can internally recover cue-level evidence from the image at inference time. We therefore ask whether ViCuR reduces the concrete grounding errors induced by answer-based OPSD, and whether the sink-based recovery module aggregates visual tokens consistent with the teacher-side cue evidence. We use two representative examples---one from Geometry3K and one from MathVista---because they stress different forms of recoverable visual evidence: local relation binding in a geometry diagram and query-conditioned evidence selection in a chart.

\begin{figure}[!htbp]
\centering
\begin{minipage}[b]{0.47\linewidth}
    \centering
    \includegraphics[width=0.50\linewidth]{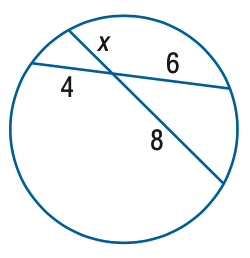}
    \vspace{0.3em}

    \colorbox{gray!8}{\parbox{0.88\linewidth}{\small
    \textbf{Q:} Find \(x\). \hfill \textbf{A:} \(3\)
    }}
    \vspace{0.2em}

    {\footnotesize (a) Case~1: Geometry3K}
\end{minipage}
\hfill
\begin{minipage}[b]{0.47\linewidth}
    \centering
    \includegraphics[width=0.72\linewidth]{imgs/visual_imgs/q2.png}
    \vspace{0.3em}

    \colorbox{gray!8}{\parbox{0.88\linewidth}{\small
    \textbf{Q:} Which model has the highest Acc score when Pretrain Loss = 1.80? \\[2pt]
    \textbf{A:} RFT U13B
    }}
    \vspace{0.2em}

    {\footnotesize (b) Case~2: MathVista}
\end{minipage}
\caption{Overview of the two cases used in the qualitative study.}
\label{fig:case_overview}
\end{figure}

\subsection{Case 1: Geometry3K --- Intersecting Chords Theorem}
\label{app:case_geo3k}

\paragraph{Quantitative stability.}
We evaluate each model on this Geometry3K example across 10 repeated runs with identical input. The base Qwen3-VL-2B-Instruct model produces the correct answer in only 5/10 runs, and the answer-based OPSD baseline improves to 7/10, while OPSD + ViCuR answers correctly in all 10/10 runs. This example isolates the intended role of visual cue recovery because the error is not about theorem recall: all models invoke the intersecting chords theorem. The decisive variable is whether the student recovers the visual cue that the diagram contains two intersecting chords, with \(x\) paired with \(8\) on one chord and \(4\) paired with \(6\) on the other. The incorrect runs instead bind \(4\) and \(x\) as if they were on the same chord, yielding \(4 \times x = 6 \times 8\) and \(x=12\). ViCuR recovers the cue-level relation needed for the theorem, giving \(4 \times 6 = x \times 8\) and \(x=3\). Thus, the improvement is better interpreted as recovery of inference-accessible visual structure rather than exposure to an answer or a reasoning shortcut.

\paragraph{Model reasoning outputs.}
Below we compare the reasoning traces of the three models. The base model and OPSD baseline produce plausible-looking derivations, but the derivations are anchored to the wrong recovered cue: they place \(4\) and \(x\) on the same chord. ViCuR changes the critical grounding step before any algebra is performed. Its trace matches the teacher-side visual cue format used in our framework---a description of visible segment relations---rather than a direct answer-bearing rationale, and the subsequent computation follows the correct diagram structure.

\medskip
\begin{tcolorbox}[enhanced, breakable, frame hidden,
    borderline west={2.5pt}{0pt}{red!70!black},
    colback=red!3, left=5pt, right=3pt, top=3pt, bottom=3pt,
    before skip=5pt, after skip=5pt]
{\bfseries\small Base Qwen3-VL-2B-Instruct {\normalfont(incorrect, $x=12$)}}

\smallskip\footnotesize
Based on the provided image, we can solve for $x$ using the **Intersecting Chords Theorem**.

This theorem states that when two chords intersect inside a circle, the product of the lengths of the segments of one chord is equal to the product of the lengths of the segments of the other chord.

In the diagram, the two chords intersect at a point inside the circle. The segments are:
- One chord is divided into segments of lengths 4 and $x$.
- The other chord is divided into segments of lengths 6 and 8.

The theorem gives us the equation:
$$4 \times x = 6 \times 8$$

Now, we solve for $x$:
$$4x = 48$$
$$x = \frac{48}{4}$$
$$x = 12$$

Therefore, the value of $x$ is 12.
\end{tcolorbox}

\begin{tcolorbox}[enhanced, breakable, frame hidden,
    borderline west={2.5pt}{0pt}{orange!80!black},
    colback=orange!3, left=5pt, right=3pt, top=3pt, bottom=3pt,
    before skip=5pt, after skip=5pt]
{\bfseries\small OPSD baseline {\normalfont(incorrect, $x=12$)}}

\smallskip\footnotesize
To find the value of $x$, we can use **the Intersecting Chords Theorem**. This theorem states that if two chords intersect inside a circle, the products of the lengths of the segments of each chord are equal.

In the given diagram, the two chords intersect at a point inside the circle. The chords are:
- One chord is divided into segments of lengths $4$ and $x$.
- The other chord is divided into segments of lengths $6$ and $8$.

According to the theorem:
$$(4)(x) = (6)(8)$$

Now, let's solve for $x$:
$$4x = 48$$
$$x = \frac{48}{4}$$
$$x = 12$$

So, the value of $x$ is $12$.
\end{tcolorbox}

\begin{tcolorbox}[enhanced, breakable, frame hidden,
    borderline west={2.5pt}{0pt}{blue!70!black},
    colback=blue!3, left=5pt, right=3pt, top=3pt, bottom=3pt,
    before skip=5pt, after skip=5pt]
{\bfseries\small OPSD + ViCuR (ours) {\normalfont(correct, $x=3$)}}

\smallskip\footnotesize
To find the value of $x$, we can use the **Intersecting Chords Theorem**. This theorem states that if two chords intersect inside a circle, the products of the lengths of the segments of one chord are equal to the products of the lengths of the segments of the other chord.

In this case, the two chords intersect at a point inside the circle. The chords are:
- One chord is divided into segments of length 4 and 6.
- The other chord is divided into segments of length $x$ and 8.

According to the Intersecting Chords Theorem:
$$(4)(6) = (x)(8)$$

Now, solve for $x$:
$$24 = 8x$$
$$x = \frac{24}{8} = 3$$

So, the value of $x$ is $\boxed{3}$.
\end{tcolorbox}

\subsection{Case 2: MathVista --- Generalization Beyond Geometry}
\label{app:case_mathvista}

\paragraph{Qualitative observations.}
This example from MathVista examines whether cue recovery extends beyond structured geometry diagrams to chart-reading tasks. As shown in Fig.~\ref{fig:case_overview}, the question asks which model achieves the highest accuracy at a given pretrain loss value. The recoverable evidence is no longer a small set of local geometric labels; it is a query-conditioned evidence set consisting of the \(x\)-axis slice at \(\mathrm{Pretrain\ Loss}=1.80\), the identities of multiple chart series, and their relative vertical positions. The base model and OPSD baseline both answer LLaMA-2-70B, suggesting that answer-based supervision does not prevent the student from relying on an incomplete or misread subset of curves. In contrast, ViCuR treats the problem as a visual comparison over the relevant slice and identifies RFT U13B as the highest. This pattern is consistent with the quantitative improvements on MathVista (e.g., +2.2 for OPSD + ViCuR at 2B scale, Tab.~\ref{tab:main_results}), while illustrating that the recovered cue evidence need not be geometry-specific.

\paragraph{Model reasoning outputs.}
Below we compare the full reasoning traces. The base model and OPSD baseline list confident numerical estimates, but their comparison omits or mispositions the series that should dominate at the queried loss value. ViCuR's trace is more faithful to the visual cue that a teacher would provide for this task: locate the specified \(x\)-coordinate, compare the series at that slice, and select the highest curve. The case therefore probes whether ViCuR recovers a task-conditioned evidence set, not merely whether it memorizes a chart answer.

\medskip
\begin{tcolorbox}[enhanced, breakable, frame hidden,
    borderline west={2.5pt}{0pt}{red!70!black},
    colback=red!3, left=5pt, right=3pt, top=3pt, bottom=3pt,
    before skip=5pt, after skip=5pt]
{\bfseries\small Base Qwen3-VL-2B-Instruct {\normalfont(incorrect: LLaMA-2-70B)}}

\smallskip\footnotesize
Based on the provided chart, we can determine which model has the highest accuracy (Acc) at a Pretrain Loss of 1.80 by examining the data points on the graph.

1.  **Locate the Pretrain Loss:** Find the value `1.80` on the horizontal axis (x-axis), labeled "Pretrain Loss".
2.  **Identify the corresponding Acc scores:** Look at the vertical axis (y-axis), labeled "Acc", to see the accuracy for each model at this point.
3.  **Compare the values:** The chart shows the following accuracy values at `Pretrain Loss = 1.80`:
    -   **LLaMA-7B:** Accuracy is approximately 37.5
    -   **LLaMA-13B:** Accuracy is approximately 42.5
    -   **LLaMA-2-7B:** Accuracy is approximately 45.5
    -   **LLaMA-2-13B:** Accuracy is approximately 47.5
    -   **LLaMA-2-70B:** Accuracy is approximately 61.5
    -   **RFT k=100:** Accuracy is approximately 40.5
    -   **RFT k=25:** Accuracy is approximately 41.5
    -   **RFT k=6:** Accuracy is approximately 42.5
    -   **RFT U13B:** Accuracy is approximately 48.5

The highest accuracy score among these models is **61.5**, which corresponds to the **LLaMA-2-70B** model.

Therefore, the model with the highest Acc score when Pretrain Loss is equal to 1.80 is **LLaMA-2-70B**.
\end{tcolorbox}

\begin{tcolorbox}[enhanced, breakable, frame hidden,
    borderline west={2.5pt}{0pt}{orange!80!black},
    colback=orange!3, left=5pt, right=3pt, top=3pt, bottom=3pt,
    before skip=5pt, after skip=5pt]
{\bfseries\small OPSD baseline {\normalfont(incorrect: LLaMA-2-70B)}}

\smallskip\footnotesize
Based on the provided chart, we can determine which model has the highest Accuracy (Acc) score at a Pretrain Loss of 1.80 by examining the data points on the graph.

1.  **Locate the Pretrain Loss:** Find the value 1.80 on the horizontal axis (Pretrain Loss).
2.  **Identify the corresponding Accuracy:** Look at the vertical positions of the data points on the chart for each model at this specific Pretrain Loss value.
3.  **Compare the scores:** The chart shows the following accuracy values at Pretrain Loss = 1.80:
    - **LLaMA-7B:** The data point is at approximately 39.0 Acc.
    - **LLaMA-13B:** The data point is at approximately 43.0 Acc.
    - **LLaMA-2-7B:** The data point is at approximately 45.0 Acc.
    - **LLaMA-2-13B:** The data point is at approximately 47.0 Acc.
    - **LLaMA-2-65B:** The data point is at approximately 50.0 Acc.
    - **LLaMA-2-70B:** The data point is at approximately 54.0 Acc.

The model with the highest accuracy at a Pretrain Loss of 1.80 is **LLaMA-2-70B**.
\end{tcolorbox}

\begin{tcolorbox}[enhanced, breakable, frame hidden,
    borderline west={2.5pt}{0pt}{blue!70!black},
    colback=blue!3, left=5pt, right=3pt, top=3pt, bottom=3pt,
    before skip=5pt, after skip=5pt]
{\bfseries\small OPSD + ViCuR (ours) {\normalfont(correct: RFT U13B)}}

\smallskip\footnotesize
Based on the provided chart, we can determine which model has the highest accuracy (Acc) at a pretrain loss of 1.80 by following these steps:

1.  **Locate the Pretrain Loss value:** Find the value `1.80` on the horizontal axis (x-axis), labeled "Pretrain Loss".
2.  **Identify the corresponding accuracy:** Look at the vertical position (y-axis, labeled "Acc") for each model's line at this specific point.
3.  **Compare the values:** The y-axis represents accuracy, with higher values indicating better performance.

Let's examine the data points for the models at `Pretrain Loss = 1.80`:
-   **ICL (black line):** The data point is at approximately `40.0` Acc.
-   **SFT (blue line):** The data point is at approximately `35.0` Acc.
-   **SFT 1/2 (dotted blue line):** The data point is at approximately `37.0` Acc.
-   **SFT 1/4 (dotted light blue line):** The data point is at approximately `25.0` Acc.
-   **SFT 1/8 (dotted light purple line):** The data point is at approximately `18.0` Acc.
-   **RFT k=100 (dark green line):** The data point is at approximately `42.0` Acc.
-   **RFT k=25 (dotted light green line):** The data point is at approximately `40.0` Acc.
-   **RFT k=6 (dotted light grey line):** The data point is at approximately `40.0` Acc.
-   **RFT U13B (teal line):** The data point is at approximately `50.0` Acc.

The highest accuracy value among all models at `Pretrain Loss = 1.80` is `50.0`, which corresponds to the **RFT U13B** model.
\end{tcolorbox}

\begin{figure}[!htbp]
\centering
\includegraphics[width=\linewidth]{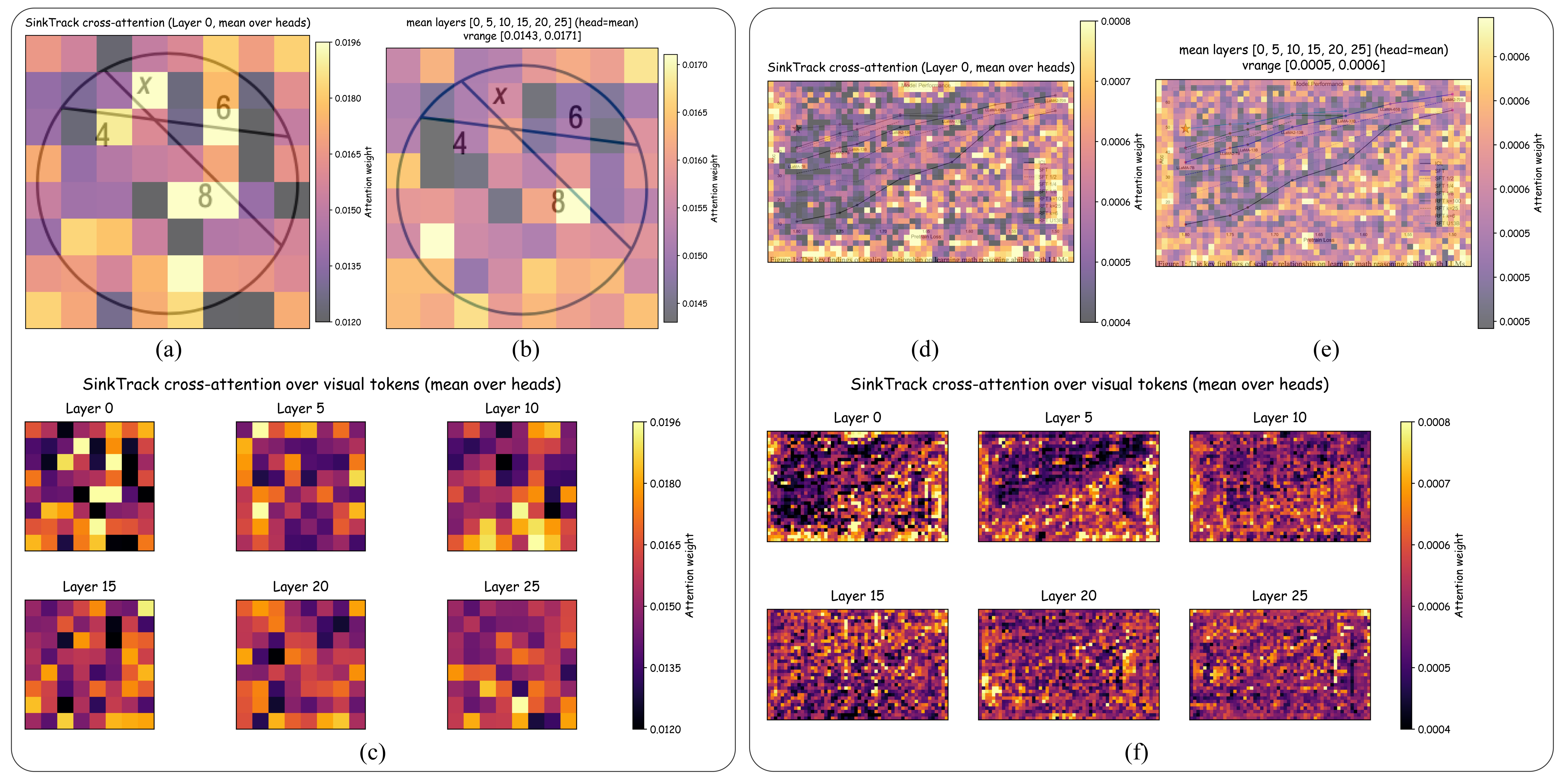}
\caption{SinkTrack cross-attention visualizations for both qualitative cases. (a)--(c) Geometry3K (Case~1): first SinkTrack layer, mean over all 6~layers, and full layer-by-layer progression. (d)--(f) MathVista (Case~2): same three views. Best viewed in color.}
\label{fig:sinktrack_combined}
\end{figure}

\paragraph{SinkTrack attention visualization.}
Fig.~\ref{fig:sinktrack_combined} presents the SinkTrack cross-attention heatmaps for both cases in a unified view. These heatmaps visualize the dedicated cross-attention from the sink token to visual tokens, i.e., the pathway through which the recovery module forms the internal representation \(\hat{S}\). They therefore test a more specific question than ordinary saliency visualization: whether the module aggregates visual evidence compatible with the teacher-side cues, despite receiving no cue text or cue-generation loss. Panels (a)--(c) correspond to the Geometry3K example (Case~1): (a)~the first SinkTrack layer, (b)~the mean attention over all 6~SinkTrack layers, and (c)~the full layer-by-layer progression. Panels (d)--(f) show the same three views for the MathVista example (Case~2).

For Case~1, the visual cue to be recovered is the chord-pairing structure: the labels \(x\) and \(8\) belong to one chord, while \(4\) and \(6\) belong to the other. In both the first-layer map and the layer-averaged map, high-attention regions overlap with the numerical labels, the intersection area, and the chord segments adjacent to these labels. The layer-by-layer views in panel~(c) further show repeated coverage of the relevant label and chord regions across SinkTrack layers, rather than a single-layer spike. This is consistent with the behavioral trace: ViCuR does not need a teacher hint that states the theorem or the answer, but it must recover the visible segment relation that makes the theorem usable.

For Case~2, the attention pattern is naturally less localized. Chart reading requires a query-conditioned aggregation over several visual elements: the \(x\)-axis value, the legends or series identities, and the vertical positions of multiple curves. Accordingly, panels~(d)--(f) show distributed, structured responses across the visual-token sequence rather than a compact hotspot over a single region. This should not be over-interpreted as proof that the model isolated the exact RFT U13B curve at every layer. A more defensible interpretation is that SinkTrack preserves multi-region chart context needed for the final comparison, which is precisely the kind of internal cue evidence that ViCuR is designed to recover.

\vspace{0.5em}
\noindent\textbf{Summary.}
Across both cases, the main advantage of ViCuR is not a change in the final reasoning template, but a more reliable conversion of inference-accessible visual cue supervision into student-side internal evidence recovery. In the geometry example, this recovery appears as correct binding between labels and chord segments; in the chart example, it appears as preserving enough multi-region context to compare series at the queried loss value. The SinkTrack visualizations do not constitute causal proof, but they provide mechanism-consistent evidence that the dedicated sink cross-attention aggregates the kind of visual information described by teacher-side cues. These qualitative observations complement the theory and main experiments by showing how ViCuR can reduce answer-privilege-induced grounding failures in concrete model outputs.


\section{Additional Theoretical Details}
\label{app:theory}

\subsection{Derivation and Analysis of Proposition 1 (Recoverable Privilege)}
\label{app:proof_prop1}

\paragraph{Proposition 1 (Recoverable Privilege).}
\textit{
Under the deterministic abstraction $S=f(X)$, visually grounded cues do not introduce the privilege-induced conditional information gap associated with answer-dependent variables. The resulting teacher supervision is conditioned only on information determined by the inference-time input $X$.
}

\medskip

In this appendix, we provide additional details on the theoretical motivation in Section~\ref{subsec:privilege_gap}. The goal is not to claim that the student can perfectly reconstruct visual cue text, but to clarify why image-grounded cues differ from answer-dependent privileges: under an ideal deterministic abstraction, the cue does not introduce information beyond the inference-time input \(X\).

\paragraph{Setup.}
Let \(X=(I,Q)\) denote the multimodal input available at inference time, and let \(S\) denote the teacher-side visual cue text. We consider a decoding step \(t\) with prefix \(Y_{<t}\). The cue-conditioned teacher distribution is
\begin{equation}
    p_T(Y_t \mid X,S,Y_{<t}),
\end{equation}
and the student distribution is
\begin{equation}
    p_\theta(Y_t \mid X,Y_{<t}).
\end{equation}
For comparison, we define the teacher's marginal distribution over \(S\) as
\begin{equation}
    p_T(Y_t \mid X,Y_{<t})
    =
    \mathbb{E}_{S \sim p(S \mid X)}
    \left[
    p_T(Y_t \mid X,S,Y_{<t})
    \right].
\end{equation}

We analyze the expected token-level KL objective
\begin{equation}
\mathcal{L}_{\mathrm{KL}}(\theta)
=
\mathbb{E}_{X,S,Y_{<t}}
\left[
D_{\mathrm{KL}}
\Big(
p_T(\cdot \mid X,S,Y_{<t})
\;\|\;
p_\theta(\cdot \mid X,Y_{<t})
\Big)
\right].
\end{equation}

\paragraph{Step 1: Expanding the KL Objective.}
Following the decomposition framework of~\cite{yang2026selfdistilledrlvr}, we expand the KL divergence as
\begin{align}
\mathcal{L}_{\mathrm{KL}}(\theta)
&=
\mathbb{E}_{X,S,Y_{<t},Y_t}
\left[
\log p_T(Y_t \mid X,S,Y_{<t})
-
\log p_\theta(Y_t \mid X,Y_{<t})
\right],
\label{eq:app_kl_expand}
\end{align}
where \(Y_t\) is sampled from \(p_T(\cdot \mid X,S,Y_{<t})\). Adding and subtracting
\(\log p_T(Y_t \mid X,Y_{<t})\), we obtain
\begin{equation}
\normalsize
\begin{aligned}
\mathcal{L}_{\mathrm{KL}}(\theta)
&=
\mathbb{E}_{X,S,Y_{<t},Y_t}
\left[
\log
\frac{
p_T(Y_t \mid X,S,Y_{<t})
}{
p_T(Y_t \mid X,Y_{<t})
}
\right] \\
&\quad+
\mathbb{E}_{X,S,Y_{<t},Y_t}
\left[
\log
\frac{
p_T(Y_t \mid X,Y_{<t})
}{
p_\theta(Y_t \mid X,Y_{<t})
}
\right].
\end{aligned}
\label{eq:app_add_subtract}
\end{equation}

\paragraph{Step 2: Information-Theoretic Decomposition.}
The first term in Eq.~\eqref{eq:app_add_subtract} is the conditional mutual information between the teacher's next token and the cue:
\begin{equation}
    I_T(Y_t;S \mid X,Y_{<t}),
\end{equation}
where the subscript \(T\) emphasizes that \(Y_t\) is drawn from the cue-conditioned teacher distribution. The second term equals
\begin{equation}
\mathbb{E}_{X,S,Y_{<t}}
\left[
\mathbb{E}_{Y_t \sim p_T(\cdot \mid X,S,Y_{<t})}
\left[
\log
\frac{
p_T(Y_t \mid X,Y_{<t})
}{
p_\theta(Y_t \mid X,Y_{<t})
}
\right]
\right].
\end{equation}
Note that in general this is \emph{not} the same as \(D_{\mathrm{KL}}(p_T(\cdot \mid X,Y_{<t}) \| p_\theta(\cdot \mid X,Y_{<t}))\), because \(Y_t\) is sampled from the cue-conditioned distribution \(p_T(\cdot \mid X,S,Y_{<t})\) rather than the marginal \(p_T(\cdot \mid X,Y_{<t})\). However, under the deterministic cue abstraction \(S=f(X)\) introduced in Step~3 below, these two distributions coincide, and the second term reduces to the standard KL between the teacher's marginal and the student:
\begin{equation}
\mathcal{L}^*(\theta)
:=
\mathbb{E}_{X,Y_{<t}}
\left[
D_{\mathrm{KL}}
\Big(
p_T(\cdot \mid X,Y_{<t})
\;\|\;
p_\theta(\cdot \mid X,Y_{<t})
\Big)
\right].
\end{equation}
Under this abstraction, we therefore obtain
\begin{equation}
\mathcal{L}_{\mathrm{KL}}(\theta)
=
\mathcal{L}^*(\theta)
+
I_T(Y_t;S \mid X,Y_{<t}).
\label{eq:app_decomposition}
\end{equation}

\paragraph{Step 3: Deterministic Cue Abstraction.}
For the purpose of theoretical analysis, we idealize the cue construction process as a deterministic mapping
\begin{equation}
    S=f(X).
\end{equation}
Under this abstraction, \(H(S\mid X)=0\). Since \(S\) is already determined once \(X\) is given, conditioning additionally on \(Y_{<t}\) or \(Y_t\) cannot introduce uncertainty about \(S\):
\begin{equation}
    H(S\mid X,Y_{<t}) = H(S\mid X,Y_{<t},Y_t)=0.
\end{equation}
Therefore,
\begin{equation}
I_T(Y_t;S \mid X,Y_{<t})
=
H(S \mid X,Y_{<t})
-
H(S \mid X,Y_{<t},Y_t)
=
0.
\end{equation}
In this idealized case, cue-conditioned supervision does not introduce an additional conditional information gap beyond \(X\), and Eq.~\eqref{eq:app_decomposition} reduces to
\begin{equation}
    \mathcal{L}_{\mathrm{KL}}(\theta)
    =
    \mathcal{L}^*(\theta).
\end{equation}

\paragraph{Discussion: Information Availability vs. Recoverability.}
The above result should be interpreted carefully. It does not imply that the student directly observes \(S\), nor that a standard model can perfectly reconstruct the cue text. It only shows that, under the deterministic cue abstraction, the teacher's additional conditioning variable is determined by the inference-time input. This differs from answer-dependent privileges, where the teacher may rely on variables not recoverable from \(X\).

Thus, replacing answer-based privilege with visual cues reframes the problem. The remaining challenge is not that the teacher uses information absent at test time, but that the student must learn to extract and represent the relevant visual evidence effectively. This motivates our attention-sink-based recovery module, which provides an internal pathway for aggregating cue-level evidence from visual tokens.

\paragraph{Remark on Stochastic Cue Construction.}
As discussed in Section~\ref{subsec:privilege_gap}, the deterministic abstraction \(S=f(X)\) is an idealized limit. In practice, cue text may be produced by an LLM-based pipeline with sampling, yielding \(H(S\mid X)>0\). In that case, the conditional information term does not vanish exactly, and the decomposition in Eq.~\eqref{eq:app_decomposition} retains a nonzero \(I_T(Y_t;S \mid X,Y_{<t})\) component. However, because visual cues describe evidence visible in \(X\) rather than answer-dependent information, this residual term is expected to be substantially smaller than for answer-based privileges where \(H(R \mid X)\) can be large. Our method does not require token-level reconstruction of \(S\); it uses the teacher's cue-conditioned behavior to shape the student's internal visual representation, making it less sensitive to superficial linguistic variation in the cue text.

\subsection{Gradient Analysis of Proposition 2 (Task-Driven Implicit Alignment)}
\label{app:analysis_prop2}

\paragraph{Proposition 2 (Task-Driven Implicit Alignment).}
\textit{
Under the advantage-weighted on-policy distillation objective, the recovery module receives nonzero gradients to the extent that its internal visual evidence representation affects student predictions on teacher-evaluated tokens. Specifically, when the advantage \(A_n > 0\) (teacher assigns higher probability than the student), gradients reinforce the current evidence representation; when \(A_n < 0\), gradients push the representation away from its current state. The distillation objective therefore implicitly encourages the sink-based representation to retain task-relevant visual evidence aligned with the teacher's visual cues, without requiring explicit cue-text generation or text-level matching.
}

\medskip

In this section, we analyze how the on-policy distillation objective provides an implicit task-driven alignment signal for the attention-sink-based recovery module. The analysis uses a simplified unclipped policy-gradient surrogate for clarity. Our implementation follows a PPO-style objective with old log-probabilities, importance ratios, and clipping, but the key gradient pathway into the recovery module is the same.

\paragraph{Setup.}
Let \(X=(I,Q)\) denote the inference-time input. We denote the internal cue-level evidence representation induced by the recovery module as
\begin{equation}
\hat S = g(V,Q;\theta_{\mathrm{sink}}),
\end{equation}
where \(V\) represents the layer-wise visual token states and \(\theta_{\mathrm{sink}}\) denotes the recovery-module parameters. The student distribution is written as
\begin{equation}
p_\theta(\hat y_n \mid X,\hat S,\hat y_{<n}),
\end{equation}
and the cue-conditioned teacher distribution is
\begin{equation}
p_T(\hat y_n \mid X,S,\hat y_{<n}).
\end{equation}

For a sampled student trajectory \(\hat y \sim p_\theta(\cdot\mid X)\), we define the teacher-shaped token advantage as
\begin{equation}
A_n(X,\hat y)
=
\log p_T(\hat y_n \mid X,S,\hat y_{<n})
-
\log p_\theta(\hat y_n \mid X,\hat S,\hat y_{<n}).
\end{equation}
During policy-gradient optimization, this advantage is treated as a stop-gradient quantity.

\paragraph{Step 1: Basic Policy-Gradient Surrogate.}
The simplified distillation surrogate is
\begin{equation}
\mathcal{L}_{\mathrm{distill}}(\theta)
=
-
\mathbb{E}_{X \sim \mathcal{D}}
\mathbb{E}_{\hat y \sim p_\theta}
\left[
\frac{1}{|\hat y|}
\sum_{n=1}^{|\hat y|}
\mathrm{sg}\!\left[A_n(X,\hat y)\right]\,
\log p_\theta(\hat y_n \mid X,\hat S,\hat y_{<n})
\right],
\end{equation}
where \(\mathrm{sg}[\cdot]\) denotes stop-gradient. Therefore,
\begin{equation}
\nabla_{\theta_{\mathrm{sink}}}\mathcal{L}_{\mathrm{distill}}
=
-
\mathbb{E}
\left[
\frac{1}{|\hat y|}
\sum_{n=1}^{|\hat y|}
\mathrm{sg}\!\left[A_n(X,\hat y)\right]\,
\nabla_{\theta_{\mathrm{sink}}}
\log p_\theta(\hat y_n \mid X,\hat S,\hat y_{<n})
\right].
\label{eq:app_pg_grad}
\end{equation}

\paragraph{Step 2: Chain Rule Through the Recovery Module.}
Since the recovery parameters influence the student distribution through \(\hat S=g(V,Q;\theta_{\mathrm{sink}})\), we have
\begin{equation}
\nabla_{\theta_{\mathrm{sink}}}
\log p_\theta(\hat y_n \mid X,\hat S,\hat y_{<n})
=
\nabla_{\hat S}
\log p_\theta(\hat y_n \mid X,\hat S,\hat y_{<n})
\cdot
\nabla_{\theta_{\mathrm{sink}}}g(V,Q;\theta_{\mathrm{sink}}).
\end{equation}
Substituting this into Eq.~\eqref{eq:app_pg_grad} yields
\begin{equation}
\normalsize
\nabla_{\theta_{\mathrm{sink}}}\mathcal{L}_{\mathrm{distill}}
= -
\mathbb{E}
\Bigg[
\frac{1}{|\hat y|}
\sum_{n=1}^{|\hat y|}
\mathrm{sg}\!\left[A_n(X,\hat y)\right]\,
\nabla_{\hat S}
\log p_\theta(\hat y_n \mid X,\hat S,\hat y_{<n})
\cdot
\nabla_{\theta_{\mathrm{sink}}}g(V,Q;\theta_{\mathrm{sink}})
\Bigg].
\label{eq:app_pg_chain}
\end{equation}

\paragraph{Interpretation.}
Equation~\eqref{eq:app_pg_chain} shows that the gradient flowing into the recovery module is modulated by the teacher-student token-level discrepancy. Specifically, when \(A_n > 0\) (the cue-conditioned teacher assigns higher probability than the student to the sampled token), the gradient reinforces the current evidence representation, encouraging the module to retain and strengthen whatever visual evidence contributed to the student's prediction. When \(A_n < 0\), the gradient pushes the representation away from its current state, discouraging evidence patterns that led to predictions the teacher did not favor.

This provides an implicit task-driven alignment signal. Instead of forcing the student to reconstruct the cue text \(S\) token by token, the objective shapes \(\hat S\) only through its effect on downstream predictions. As a result, the recovery module is encouraged to encode visual evidence that matters for matching the teacher's cue-conditioned behavior, while avoiding unnecessary capacity spent on superficial or task-irrelevant linguistic details of the cue text.

\paragraph{Limitations of the analysis.}
The gradient derivation above uses a simplified unclipped surrogate. In practice, PPO-style clipping limits the magnitude of updates when the importance ratio deviates significantly from 1, which can dampen the alignment signal for tokens with very large advantage magnitudes. Additionally, the analysis treats \(\hat{S}\) as a single vector for clarity, whereas in practice the recovery module updates the sink token at multiple layers, and the gradient pathway involves interactions across layers. These simplifications do not affect the qualitative conclusion---that the distillation objective provides a task-driven signal for the recovery module---but the precise dynamics may differ from the idealized single-vector picture.

\subsection{Analytical Counterpart: The Full-Vocabulary KL Objective}
\label{app:kl_counterpart}

Although our implementation optimizes a sampled-token on-policy objective, it is useful to consider the full-vocabulary forward KL objective as an analytical counterpart. This is the same divergence direction ($D_{\mathrm{KL}}(p_T \| p_\theta)$) used in OPSD~\cite{OPSD}, where the teacher's token-wise distribution serves as the reference. Analyzing this objective connects our method with standard knowledge distillation and clarifies how cue-conditioned teacher distributions can shape the student's internal visual evidence representation.

\paragraph{Setup.}
Let \(X=(I,Q)\) denote the inference-time input and \(S\) the teacher-side visual cue text. Let
\begin{equation}
\hat S = g(V,Q;\theta_{\mathrm{sink}})
\end{equation}
be the internal representation induced by the recovery module. We write the student distribution as \(p_\theta(Y_t \mid X,\hat S,Y_{<t})\), and the cue-conditioned teacher distribution as \(p_T(Y_t \mid X,S,Y_{<t})\).

Consider the full-vocabulary distillation objective
\begin{equation}
\mathcal{L}_{\mathrm{KL}}(\theta)
=
\mathbb{E}_{X,S,Y_{<t}}
\left[
D_{\mathrm{KL}}
\Big(
p_T(\cdot \mid X,S,Y_{<t})
\;\|\;
p_\theta(\cdot \mid X,\hat S,Y_{<t})
\Big)
\right].
\end{equation}

\paragraph{Step 1: Expanding the KL Objective.}
Expanding over the vocabulary \(\mathcal{V}\) gives
\begin{equation}
\begin{aligned}
\mathcal{L}_{\mathrm{KL}}(\theta)
&=
\mathbb{E}_{X,S,Y_{<t}}
\left[
\sum_{v \in \mathcal V}
p_T(v \mid X,S,Y_{<t})
\log
\frac{
p_T(v \mid X,S,Y_{<t})
}{
p_\theta(v \mid X,\hat S,Y_{<t})
}
\right] \\
&=
C
-
\mathbb{E}_{X,S,Y_{<t}}
\left[
\sum_{v \in \mathcal V}
p_T(v \mid X,S,Y_{<t})
\log p_\theta(v \mid X,\hat S,Y_{<t})
\right].
\end{aligned}
\end{equation}
where \(C = \mathbb{E}_{X,S,Y_{<t}}[\sum_{v} p_T(v \mid X,S,Y_{<t}) \log p_T(v \mid X,S,Y_{<t})]\) denotes the teacher's conditional entropy and is constant with respect to the student parameters \(\theta\) (assuming \(Y_{<t}\) is held fixed for the purpose of this per-step analysis).

\paragraph{Step 2: Gradient Flow into the Internal Representation.}
Differentiating with respect to \(\hat S\) gives
\begin{equation}
\nabla_{\hat S}\mathcal{L}_{\mathrm{KL}}
=
-
\mathbb{E}_{X,S,Y_{<t}}
\left[
\sum_{v \in \mathcal V}
p_T(v \mid X,S,Y_{<t})
\nabla_{\hat S}
\log p_\theta(v \mid X,\hat S,Y_{<t})
\right].
\end{equation}

\paragraph{Step 3: Chain Rule to Recovery Parameters.}
Applying the chain rule through \(\hat S=g(V,Q;\theta_{\mathrm{sink}})\), we obtain
\begin{equation}
\normalsize
\nabla_{\theta_{\mathrm{sink}}}\mathcal{L}_{\mathrm{KL}}
= -
\mathbb{E}_{X,S,Y_{<t}}
\Bigg[
\sum_{v \in \mathcal V}
p_T(v \mid X,S,Y_{<t})
\nabla_{\hat S}
\log p_\theta(v \mid X,\hat S,Y_{<t})
\cdot
\nabla_{\theta_{\mathrm{sink}}}g(V,Q;\theta_{\mathrm{sink}})
\Bigg].
\label{eq:app_kl_chain}
\end{equation}

\paragraph{Interpretation.}
Equation~\eqref{eq:app_kl_chain} shows that the recovery module is updated according to a teacher-weighted combination of student score gradients. Tokens that the cue-conditioned teacher considers more likely contribute more strongly to the gradient. Therefore, the teacher distribution acts as a behavioral guide for shaping the internal representation \(\hat S\). This provides an implicit alignment signal: the student is not asked to reproduce the cue text itself, but to adjust its internal visual evidence representation so that its predictions better match the cue-conditioned teacher behavior.

\clearpage
\twocolumn



\end{document}